\documentclass[runningheads]{llncs}

\usepackage{eccv}

\usepackage{eccvabbrv}

\usepackage{graphicx}
\usepackage{amsmath}
\usepackage{amssymb}
\usepackage{booktabs}
\usepackage{multirow}
\usepackage{tikz}
\usepackage{hyperref}

\usepackage{orcidlink}

\usepackage{color, colortbl}
\definecolor{LightGray}{rgb}{0.92,0.92,0.92}
\definecolor{Gray1}{rgb}{0.95,0.95,0.95}
\definecolor{Gray2}{rgb}{0.9,0.9,0.9}
\definecolor{darkblue}{rgb}{0,0.592,0.655}%
\definecolor{darkyellow}{rgb}{0.655,0.608,0}%
\definecolor{darkgreen}{rgb}{0.416,0.659,0.310}%

\usepackage{pifont}
\newcommand{\modelname}{\emph{Idea2Img}\xspace}
\newcommand{\lmmname}{GPT-4V\xspace}
\newcommand{\lmmnamefull}{GPT-4V(ision)\xspace}
\newcommand{\idea}{\textit{IDEA}\xspace}

\usepackage[accsupp]{axessibility}  %

\usepackage[most]{tcolorbox}
\newtcolorbox{myquote}{
    breakable,
    width=0.9\textwidth,
    colback=white,
    colframe=black,
    fontupper=\itshape,
    boxrule=0.2mm,
    left=1mm,
    right=1mm,
    arc=3mm,
    auto outer arc
}

\newcommand*\circled[1]{\tikz[baseline=(char.base)]{
            \node[shape=circle,draw,inner sep=2pt] (char) {#1};}}

\makeatletter
\DeclareRobustCommand\onedot{\futurelet\@let@token\@onedot}
\def\@onedot{\ifx\@let@token.\else.\null\fi\xspace}

\def\eg{\emph{e.g}\onedot} \def\Eg{\emph{E.g}\onedot}
\def\ie{\emph{i.e}\onedot} 
 
\def\etc{\emph{etc}\onedot} \def\vs{\emph{vs}\onedot}

\usepackage[capitalize]{cleveref}
\crefname{section}{Sec.}{Secs.}
\Crefname{section}{Section}{Sections}
\Crefname{table}{Table}{Tables}
\crefname{table}{Tab.}{Tabs.}

\begin{document}

\title{\modelname: Iterative Self-Refinement with \lmmname \\ for Automatic Image Design and Generation} 

\titlerunning{\modelname: Self-Refinement with LMMs for Automatic Visual Creation}

\author{Zhengyuan Yang\orcidlink{0000-0002-5808-0889} \and
Jianfeng Wang\orcidlink{0000-0002-3156-4429} \and
Linjie Li \and
Kevin Lin \and
Chung-Ching Lin\orcidlink{0009-0003-6507-3657} \and
Zicheng Liu\orcidlink{0000-0001-5894-7828} \and
Lijuan Wang\orcidlink{0000-0002-5705-876X}}

\authorrunning{Z.~Yang et al.}

\institute{Microsoft
{\tt\footnotesize \{zhengyang,jianfw,lindsey.li,keli,chungching.lin,zliu,lijuanw\}@microsoft.com}\\
\url{https://idea2img.github.io/}}

\maketitle

\begin{abstract}

We introduce ``Idea to Image,''\footnote{Short for ``\modelname.'' System logo design~\raisebox{-3pt}{\includegraphics[height=15pt]{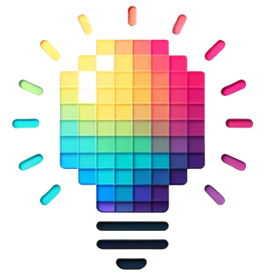}}~assisted by \modelname.} an agent system that enables multimodal iterative self-refinement with \lmmnamefull for automatic image design and generation. Humans can quickly identify the characteristics of different text-to-image (T2I) models via iterative explorations. This enables them to efficiently convert their high-level generation ideas into effective T2I prompts that can produce good images. We investigate if systems based on large multimodal models (LMMs) can develop analogous multimodal self-refinement abilities that enable exploring unknown models or environments via self-refining tries. \modelname cyclically generates revised T2I prompts to synthesize draft images, and provides directional feedback for prompt revision, both conditioned on its memory of the probed T2I model's characteristics. The iterative self-refinement brings \modelname various advantages over vanilla T2I models. Notably, \modelname can process input ideas with interleaved image-text sequences, follow ideas with design instructions, and generate images of better semantic and visual qualities. The user preference study validates the efficacy of \modelname on automatic image design and generation via multimodal iterative self-refinement.

\keywords{Multimodal Agents \and Self-Refinement \and Large Multimodal Models \and Image Design and Generation}
\end{abstract}

\section{Introduction}
\begin{figure}[t]
\centering
\includegraphics[width=1.0\textwidth]{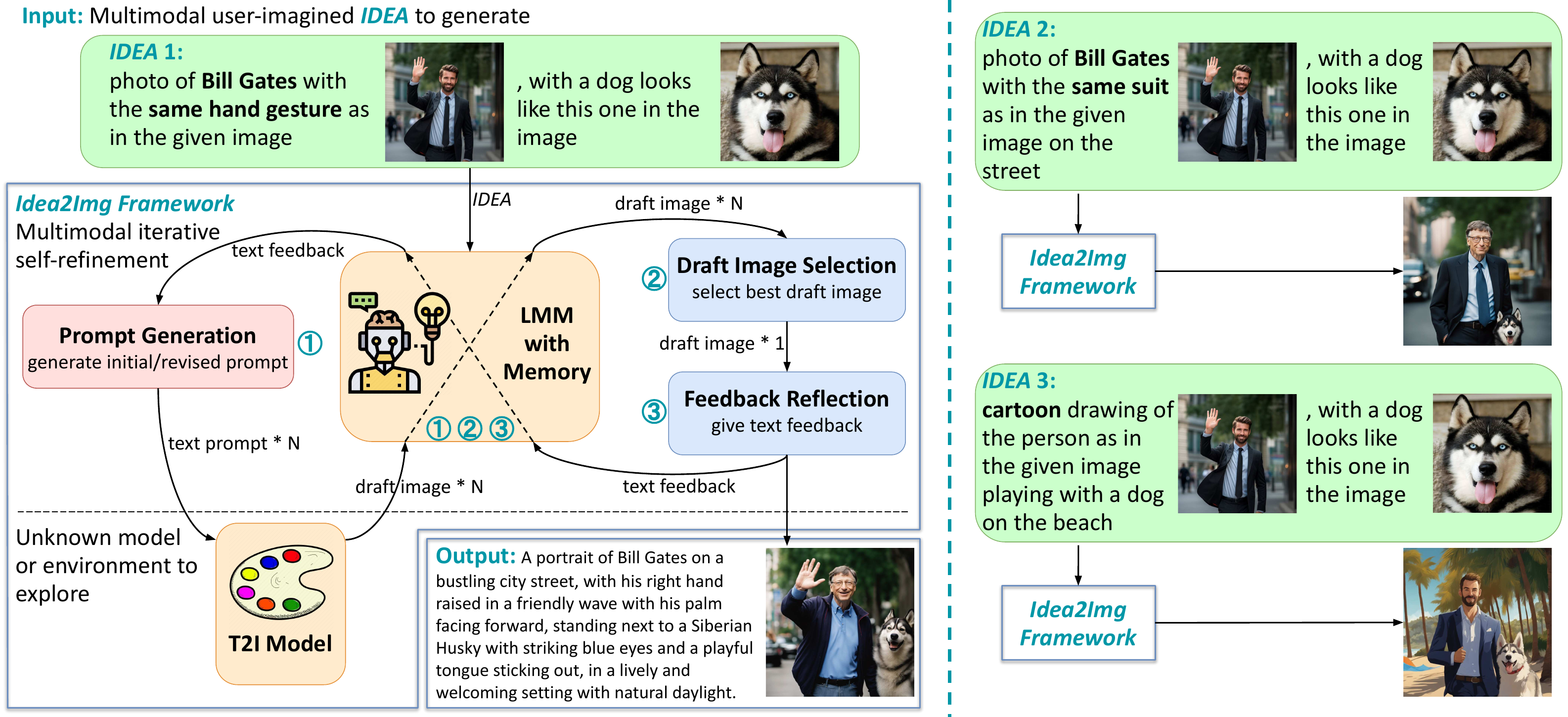}
\caption[Caption for LOF]{
    \modelname framework enables LMMs to mimic human-like exploration to use a T2I model, enabling the design and generation of an imagined image specified as a multimodal input {\color{darkblue} \textbf{{\idea}}}.
    The iterative process involves LMMs functioning in different roles to refine the image creation. Specifically, LMMs will \textbf{(1)} generate and revise text prompts for the T2I model, \textbf{(2)} select the best draft images, and \textbf{(3)} provide feedback on the errors and revision directions. This multimodal iterative self-refinement process requires LMMs to memorize the T2I model's characteristics observed in previous iterations as humans and adjust T2I prompts accordingly.
	}
\label{fig:intro}
\end{figure}

\begin{figure*}[t!]
\centering
\includegraphics[width=1.0\textwidth]{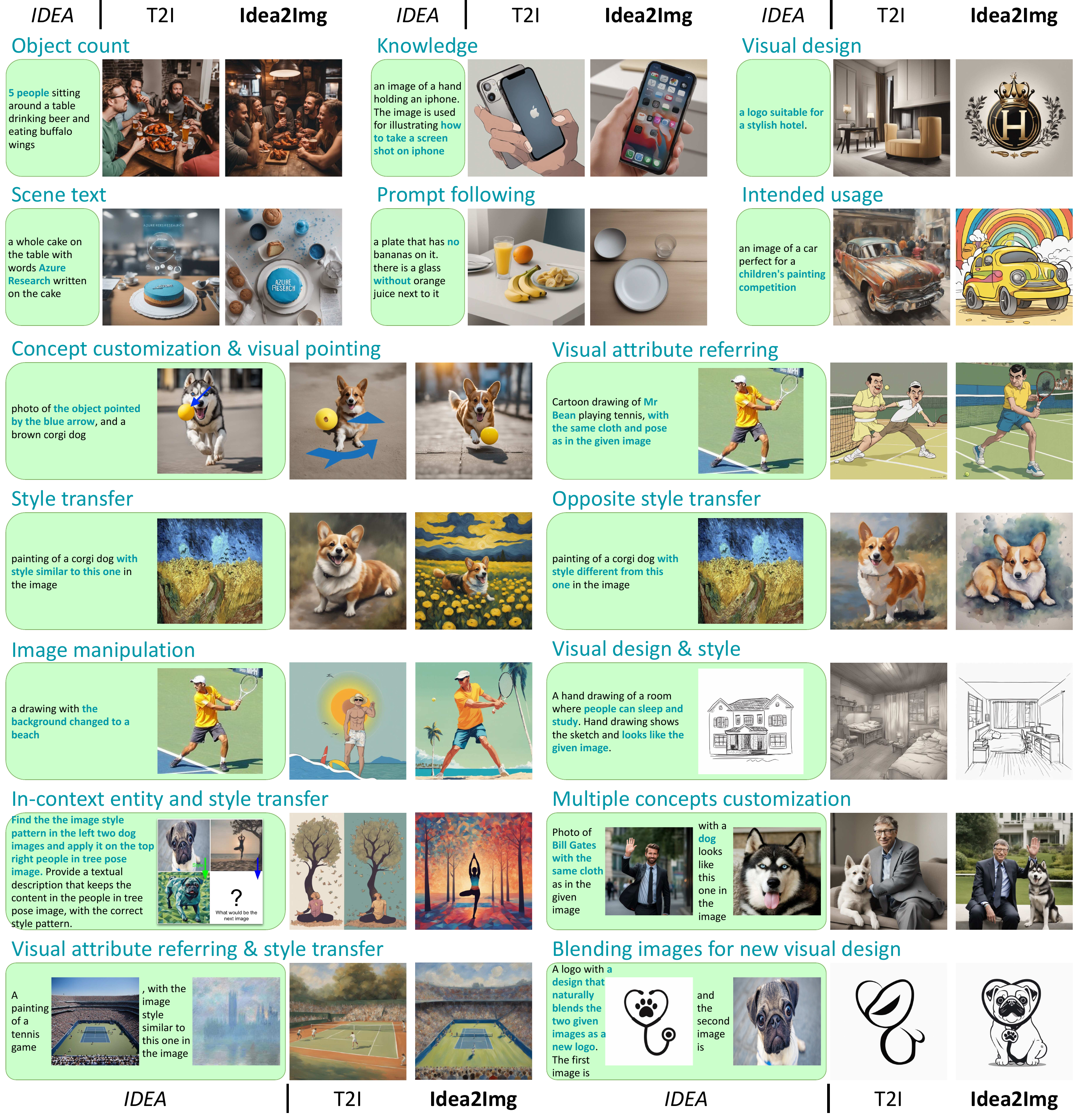}
\caption[Caption for LOF]{
    Overview of the image design and generation scenarios enabled by \modelname. In each sub-figure, the image and text in the left green box are the user input \idea. The center image is the baseline results directly generated by the same T2I model with a human-written T2I prompt, and the image on the right is generated with the T2I prompt discovered by \modelname's iterative self-refinement exploration.
	}
\label{fig:teaser}
\end{figure*}

``Image design and generation'' aims to create an image from a high-level user idea. This input {\color{darkblue} \textbf{{\idea}}} can contain interleaved reference images, such as ``the dog looks like the one in the image,'' or with instructional texts specifying the intended design usage, such as ``a logo for the Idea2Img system.'' 
To convert \idea into an image, humans may first draft detailed descriptions of the imagined image, and then use text-to-image (T2I) models~\cite{ramesh2022hierarchical,saharia2022photorealistic,yu2022scaling,rombach2022high,podell2023sdxl} to generate the image. This manual process for users to search for an ideal detailed description (\ie, T2I prompt) that fits the T2I model typically involves iterative exploration~\cite{wang2022diffusiondb,zhu2023collaborative}.
As shown in Figure~\ref{fig:intro}, humans may first design and draft an initial T2I prompt based on their imagined \idea to generate. 
Then, they can obtain multiple draft images with a T2I model, select the most promising draft, write text feedback, and further revise the T2I prompt. As this iteration progresses, we humans can swiftly grasp the characteristics of a specific T2I model, such as words that the model can not understand, finally producing a good image generated by a suitable T2I prompt. Given the remarkable capabilities of large multimodal models (LMMs)~\cite{openai2023gpt4,bard,yang2023dawn}, we explore if we can build systems based on LMMs to develop similar iterative self-refinement ability, thereby relieving humans from the tedious process of converting ideas to images. %

Iterative self-refinement is one intrinsic ability humans possess when exploring unknown environments and solving complicated problems. Large language models (LLMs) agent systems~\cite{madaan2023self,shinn2023reflexion,chen2023teaching} have demonstrated the effectiveness of self-refinement in better addressing natural language processing tasks, such as acronym generation, sentiment retrieval, text-based environment exploration, \etc. %
Transitioning from text-only tasks to multimodal environments poses new challenges of improving, assessing, and verifying multimodal contents, such as multiple interleaved image-text sequences. 
For example, when learning to use T2I models, LMMs need to improve the generation with revised T2I prompts, assess multiple images in detail to select the best draft, and verify the draft image with the multimodal \idea to provide text feedback. These steps, each requiring different multimodal understanding capabilities, jointly enable the intriguing multimodal iterative self-refinement ability.
Such an LMM framework can automatically learn to tackle various real-world problems~\cite{yang2023dawn} via self-exploration, such as navigating GUI to use electronic devices, exploring unknown physical environments via an embodied agent, engaging in electronic games, and so on. In this study, we focus on ``image design and generation'' as the task to study the multimodal iterative self-refinement ability.

To this end, we introduce \modelname, a multimodal iterative self-refinement framework for automatic image design and generation. As illustrated in Figure~\ref{fig:intro}, \modelname involves an LMM, \lmmnamefull~\cite{openai2023gpt4,gpt4v,gpt4vcontribution,gpt4vblog}, interacting with a T2I model to probe its usage and find an effective T2I prompt. The LMM will act in different roles to analyze the return signal from the T2I model (\ie, draft images) and design the next round's queries (\ie, text T2I prompts). The three roles of generating T2I prompts, selecting draft images, and reflecting feedback together enable the multimodal iterative self-refinement ability. Specifically, \textbf{(1)} Prompt generation: \lmmname~generates $N$ text prompts that correspond to the input multimodal user \idea, conditioned on the previous text feedback and refinement history; \textbf{(2)} Draft image selection: \lmmname~carefully compares $N$ draft images for the same \idea and select the most promising one; \textbf{(3)} Feedback reflection: \lmmname examines the discrepancy between the draft image and the \idea. \lmmname then provides feedback on what is incorrect, the plausible causes, and how T2I prompts may be revised to obtain a better image. Furthermore, \modelname is enhanced with a memory module that stores all prompt exploration histories, including previous draft images, text prompts, and feedback. The \modelname framework iterates among these three steps with \lmmname for automatic image design and generation.

To users, \modelname functions as an enhanced image design and generation assistant. Compared with T2I models, \modelname can handle design instructions instead of requiring detailed image description, support the multimodal \idea input, and generate images of better semantic and visual qualities. We overview representative image design and generation scenarios in Figure~\ref{fig:teaser}. %
For example, \modelname can incorporate the visual design and intended usage description in \idea, extract arbitrary visual information from the input image, and process \idea with arbitrarily interleaved image-text sequences. 
Built upon these new functionalities and scenarios of interest, we develop an evaluation \idea set with $104$ samples, containing complicated queries that humans may fail in their first trials. We perform user preference studies on \modelname with different T2I models. The consistent user preference score improvements on different image generation models, \eg, $+26.9\%$ with SDXL~\cite{podell2023sdxl}, indicate the effectiveness of \modelname in image design and generation.%

Our contributions are summarized as follows.
\begin{itemize}
    \item We study ``automatic image design and generation,'' which aims to create an image from an input \idea. This new multimodal \idea input enables visual creation with reference image inputs and instructions on desired designs. %
    \item We explore the multimodal iterative self-refinement ability in \lmmname-based agent systems, showcasing its effectiveness in improving, assessing, and verifying multimodal contents.
    \item We propose \modelname, a multimodal iterative self-refinement framework that enhances any image generation model for visual design, enabling various new image creation functionalities, and achieving better generation qualities.
    \item We present an evaluation set with $104$ challenging multimodal \idea. The consistent user preference score  improvements, when experimented on different image generation models, indicate \modelname's effectiveness in automatic image design and generation.
\end{itemize}

\section{Related Work}
\noindent\textbf{LLM-based self-refinement.}
\modelname is inspired by the effectiveness of iterative self-refinement in LLM-based agent systems~\cite{madaan2023self,shinn2023reflexion,pan2023automatically} in exploring unknown environments and tasks, built upon the successful LLM agents~\cite{yao2022react,schick2023toolformer,paranjape2023art,pryzant2023automatic,guo2023learning,zhao2023expel,yang2023large}. Self-refine~\cite{madaan2023self} takes the same LLM to iteratively critique its outputs and leverage this feedback to enhance its predictions, showing effectiveness across various NLP tasks. Reflexion~\cite{shinn2023reflexion} explores a self-reflective LLM system on the text-based environment exploration task~\cite{shridhar2020alfworld} and multi-hop QA~\cite{yang2018hotpotqa}.
Despite the success, LLM-based self-refinement naturally can not understand multimodal inputs. Consequently, the explored tasks and environments are limited to the natural language description, such as AlfWorld~\cite{shridhar2020alfworld}. \modelname explores the potential of an LMM-based iterative self-refinement system for multimodal environment exploration, from a simple T2I model to other more complicated environments.

\noindent\textbf{Multimodal agents.}
Our \modelname is related to multimodal agents~\cite{gupta2023visual,suris2023vipergpt,wu2023visual,yang2023mmreact,shen2023hugginggpt,lu2023chameleon,yu2023mm,li2023multimodal} that chain external tools such as T2I or vision-language models with LLMs for multimodal tasks. For instance, MM-ReAct~\cite{yang2023mmreact} integrates ChatGPT with multiple vision tools for multimodal reasoning and action, enabling it to solve various complicated visual understanding tasks. Visual ChatGPT~\cite{wu2023visual} empowers ChatGPT to allocate various image generation models, such as Stable Diffusion~\cite{rombach2022high}, img2img model~\cite{meng2021sdedit}, ControlNet~\cite{zhang2023adding}, enabling multi-step visual editing and generation.
The primary difference between \modelname and existing multimodal agent studies~\cite{wu2023visual,yang2023mmreact} lies in the approach to understand the tool usage. Existing studies assume the knowledge of how to best use each tool and provide such information to LLMs via text instructions or in-context examples. In contrast, the optimal usage of the tool remains unknown in \modelname and requires iterative exploration. Another minor distinction is that \modelname utilizes LMMs instead of LLMs, thereby does not require general visual understanding tools such as a caption model~\cite{wang2022git,wu2022grit}.

\noindent\textbf{Extensions of base T2I models.}
\modelname provides a more natural way for users to design and produce their desired visual content. This framework, which extends T2I models for new functionalities, is related to various works in improving base T2I models~\cite{rombach2022high,ramesh2022hierarchical,yu2022scaling,saharia2022photorealistic,podell2023sdxl}. These studies include extending the base T2I model to better follow user prompts~\cite{feng2022training,chefer2023attend,black2023training,fan2023dpok}, finding magic words in T2I prompts for better visual quality~\cite{wang2022diffusiondb,zhu2023collaborative}, supporting extra image input for image manipulation~\cite{meng2021sdedit,hertz2022prompt,brooks2023instructpix2pix,kawar2023imagic}, style transfer~\cite{gatys2015neural}, visual concept customization~\cite{ruiz2023dreambooth,kumari2023multi,avrahami2023break,chen2023subject,shi2023instantbooth}, and so on.
While specialized T2I extensions can address a single specific functionality, \modelname offers a more unified and widely applicable framework. That is, a single \modelname framework can handle various generation scenarios, ranging from style transfer to attribute customization, without requiring separate models or task-specific model design and finetune. More importantly, \modelname effectively collaborates with those enhanced generative models, consistently improving them by exploring suitable text prompts.  %

\begin{figure*}[t]
\centering
\includegraphics[width=1.\textwidth]{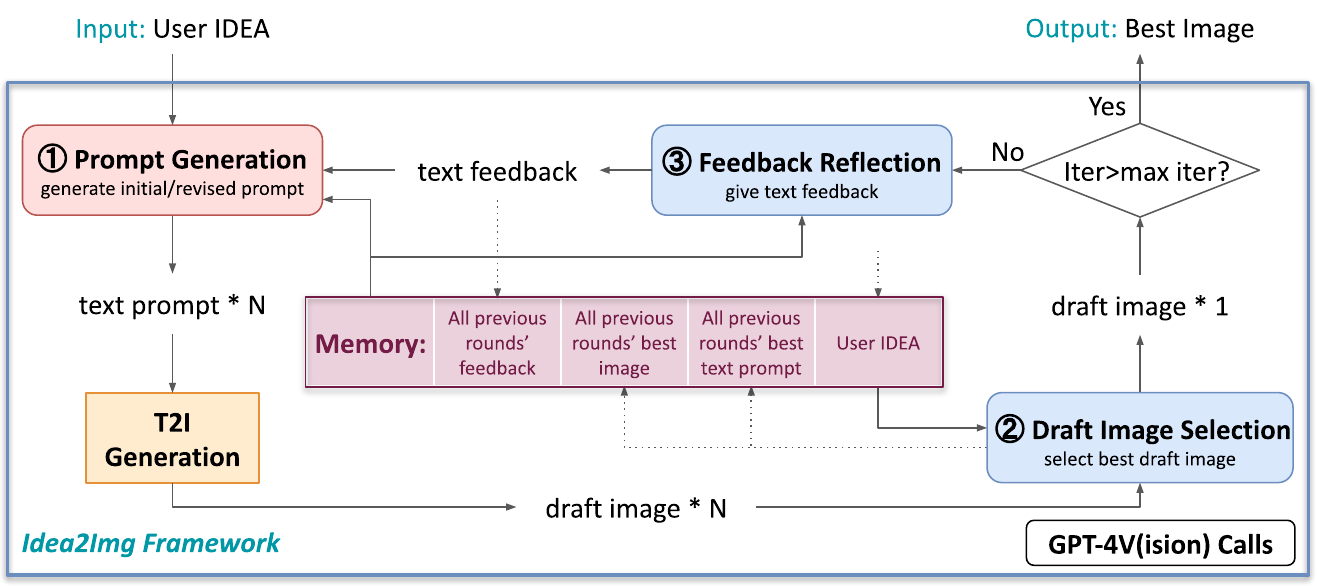}
\caption[Caption for LOF]{
    The framework overview of \modelname, which takes an LMM~\cite{openai2023gpt4,gpt4v} to explore a T2I model via multimodal iterative self-refinement, leading to an effective T2I prompt for the input user \idea. The rounded rectangle shape indicates a \lmmname call.
	}
\label{fig:arch}
\end{figure*}

\section{\modelname~Framework}
Figure~\ref{fig:arch} illustrates the \modelname framework. \modelname framework involves two core pre-trained models, \ie, the \lmmnamefull as the LMM $\mathcal{M}$ and a text-conditioned image generation model\footnote{We will show image generation models other than T2I later in experiments. For clarity, we use T2I as a representative generation model to introduce \modelname.} to explore $\mathcal{G}$. \modelname also contains a memory $m$ that stores insights on $\mathcal{G}$ discovered by $\mathcal{M}$ during previous iterations.

\noindent\textbf{Execution flow.}
We begin with an overview of the key steps in $\mathcal{M}$ iteratively exploring the use of $\mathcal{G}$. Starting from the top-left of Figure~\ref{fig:arch}, ``initial prompt generation'' converts the input multimodal user \idea into T2I text prompts, later producing multiple draft images with T2I model $\mathcal{G}$. ``Draft image selection'' then selects the best draft image among them for the current iteration. The selected image is either output as the final prediction or continues for further refinement, depending on the stop condition. For the latter, ``feedback reflection'' compares the current best draft image with the multimodal \idea, and summarizes the major discrepancy as text feedback. With the iteration history and text feedback, ``revised prompt generation'' then drafts revised T2I prompts and continues the iterative self-refinement with the new set of draft images.

\noindent\textbf{\circled{1} Initial prompt generation.}
This step generates $N$ initial T2I prompts $\left\{ y_0^0 ,\ldots, y_0^{N-1}\right\}$ following the input user \idea $x$, by prompting $\mathcal{M}$ with LMM prompt $p_{gen}$:
\begin{equation}
\label{equ:init_gen}
   \left\{ y_0^0 ,\ldots, y_0^{N-1}\right\}= \mathcal{M}(x,p_{gen})%
\end{equation}
The ``initial prompt generation'' requires $\mathcal{M}$ to understand multimodal user \idea $x$ and convert design \idea into descriptive T2I prompts. LMM prompt $p_{gen}$ is a zero-shot prompt without in-context examples. %

With the ``initial prompt generation'' step, \modelname can understand user \idea with interleaved image-text sequences, instead of the text-only T2I prompts containing the image description. Specifically, \textbf{(1)} \idea can be a high-level design or usage instead of the detailed image description, such as ``a car image for a children's painting competition''; and \textbf{(2)} \idea can take multiple images and use interleaved text instruction to extract arbitrary visual information of interest, including image style, visual entity, object attributes, \etc.
Then, in iteration $t=0$ as well as future iterations $t=t$, each T2I prompt $y_t^n$ is separately sent to the T2I model $\mathcal{G}$, resulting in $N$ draft images $i_t^n = \mathcal{G}(y_t^n), n=0,\ldots,N-1$. 

\noindent\textbf{\circled{2} Draft image selection.}
With the $N$ draft images in iteration $t$, ``draft image selection'' selects the best draft image $i_t^*$ by prompting $\mathcal{M}$ with LMM prompt $p_{select}$:
\begin{equation}
\label{equ:select}
    i_t^* = \mathcal{M}(i_t^0,\ldots,i_t^{N-1},x,p_{select}).
\end{equation}
The design of a ``draft image selection'' step is motivated by the observation that T2I models could generate bad images with good prompts. This step is designed to filter out low-quality images, and avoid the quality perturbation to dominate the iterative refinement.

The task of selecting the best image requires $\mathcal{M}$ to compare and grade both the semantics and visual quality of $N$ similar draft images. We find such a ``spot the difference'' task challenging for LMMs, and only the very recent models~\cite{openai2023gpt4,yang2023dawn} are capable of performing the selection reliably. %

\noindent\textbf{\circled{3} Feedback reflection.}
After obtaining the selected image $i_t^*$, the framework checks the stop condition, such as if the current iteration $t$ exceeds the maximum $T$. \modelname then outputs $i_t^*$ as the output image or proceeds the refinement process to the ``feedback reflection'' step accordingly. 

``Feedback reflection'' aims to provide text feedback $f_t$ that describes the direction to improve for draft image $i_t^*$. The steps prompts $\mathcal{M}$ with LMM prompt $p_{fb}$, conditioned on the draft image $i_t^*$ and memory $m$:
\begin{equation}
\label{equ:feedback}
    f_t = \mathcal{M}(i_t^*,m,x,p_{fb}).
\end{equation}
``Feedback reflection'' takes $\mathcal{M}$ to compare an image $i_t^*$ with the multimodal user \idea $x$, and summarize the gap as text feedback $f_t$. The step not only requires $\mathcal{M}$ to identify the discrepancy between image $i_t^*$ and \idea $x$, but also benefits from writing the major errors to make the iteration effective. In practice, we find it helpful to explicitly specify the aspects to check, such as style, entity, attributes, appearance, \etc, via text instructions or in-context examples in LMM prompt $p_{fb}$. Furthermore, we add text instructions to $p_{fb}$ to have $\mathcal{M}$ ``focus on one thing to improve in each feedback,'' and ``provide a high-level explanation of how to modify prompts to address the given feedback.'' %

\noindent\textbf{\circled{4}/\circled{1} Revised prompt generation.}
Finally, ``prompt generation'' takes text feedback $f_t$ and memory $m$ to draft $N$ revised prompt $\left\{ y_{t+1}^0 ,\ldots, y_{t+1}^{N-1}\right\}$, by prompting $\mathcal{M}$ with LMM prompt $p_{revise}$:
\begin{equation}
\label{equ:revise}
    \left\{ y_{t+1}^0 ,\ldots, y_{t+1}^{N-1}\right\} = \mathcal{M}(f_t,m,x,p_{revise}).
\end{equation}
Generating revised prompts requires $\mathcal{M}$ to understand the property of $\mathcal{G}$ stored in memory $m$, thereby drafting new T2I prompts that could most likely address the issue identified in $f_t$. We empirically demonstrate that \modelname can generate better prompts for $\mathcal{G}$ via iterative self-refinement. %

\noindent\textbf{Memory module.}
Memory $m$ is one important design in \modelname. $m$ has the format of interleaved image-text sequences that store all previous iterations' feedback, selected draft image, and the corresponding text prompts:
\begin{equation}
\label{equ:memory}
    m_{t} = \left[y_0^*,i_0^*,f_0, \ldots, y_{t-1}^*,i_{t-1}^*,f_{t-1}\right].
\end{equation}
It allows LMM $\mathcal{M}$ to understand the properties and capabilities of the T2I model $\mathcal{G}$ in use, such as a keyword that $\mathcal{G}$ may not understand or a complicated scene that $\mathcal{G}$ fail to generate, and incorporate such knowledge in generating the revised T2I prompts $y$. For example, it may describe the appearance of a yoga pose in detail, instead of only mentioning its name in $y$. Examples are shown in Appendix Figures~\textcolor{red}{A}-\textcolor{red}{D}, when comparing initial and refined prompts $y_0$ and $y_T$.

\section{Experiments}
\subsection{Experiment Settings}

\noindent\textbf{Compared model variants.}
We mainly compare the following three models in image generation.
\begin{itemize}
    \item ``\emph{Initial-round manual prompt}'' is the baseline T2I prompt written by humans with minor prompt engineering. It serves as the baseline of a T2I prompt that merely contains key information in \idea.
    \item ``\emph{Initial-round \modelname prompt}'' is the LMM-generated T2I prompt in the initial round. Specifically, the max iteration $T=1$, and LMM $\mathcal{M}$ is only used for initial prompt generation and draft image selection, but not feedback reflection nor revised prompt generation. This \modelname variant is used to ablate \modelname's gain from prompt generation and selection, \vs the further iterative refinement.
    \item ``\emph{Iterative self-refined \modelname prompt}'' is complete \modelname pipeline with the max iteration $T=3$.
\end{itemize}

\noindent\textbf{Evaluation samples and metrics.}
For the quantitative evaluation, we collect a dataset of $104$ user \idea as input queries. Among them, $33$ queries contain text only, $43$ queries contain an image-text sequence with a single image, and the remaining $28$ contains a sequence with two or more images. The text in most \idea contains not only descriptive content text that describes the scene to generate, but also instructional text such as ``a logo for commercial advertising'' or ``generate the pointed dog in blue.'' All test queries are manually composed.

We then perform the user preference study as the main quantitative metric. Users are presented with the \idea and multiple images to select the best one for each \idea. The evaluation script automatically shuffles the order during evaluation to prevent the influence of image orders. %
\begin{table*}[t]\small
\centering
\caption{User preference scores when applying \modelname onto different image generation models (compare the three scores in the middle section within each row individually). We observe that ``Iterative self-refined \modelname prompt'' is consistently favored across all experimented image generation models. $\Delta_\text{iteration}$ reports the preference gain from the iterative \modelname over the initial-round \modelname.}
\begin{tabular}{ l | c c c | c }
    \hline
    User preference & Initial-round & Initial-round & Iterative self-refined & \multirow{2}{*}{$\Delta_\text{iteration}$} \\
    score (\%) & manual prompt & \modelname prompt & \modelname prompt & \\
    \hline
    SDXL v1.0 & 13.5 & 29.8 & \textbf{56.7} & +26.9 \\
    DeepFloyd IF & 14.4 & 34.6 & \textbf{51.0} & +16.3 \\
    SD v2.1 & 13.5 & 40.4 & \textbf{46.2} & +5.8 \\
    SD v1.5 & 8.6 & 43.3 & \textbf{48.1} & +4.8 \\
    SDXL-img2img & 8.6 & 34.6 & \textbf{56.7} & +16.3 \\
    IF-img2img & 8.6 & 38.5 & \textbf{52.9} & +14.4 \\
    \hline
\end{tabular}
\label{table:user}
\end{table*}

\begin{figure}[t]
\centering
\includegraphics[width=0.6\textwidth]{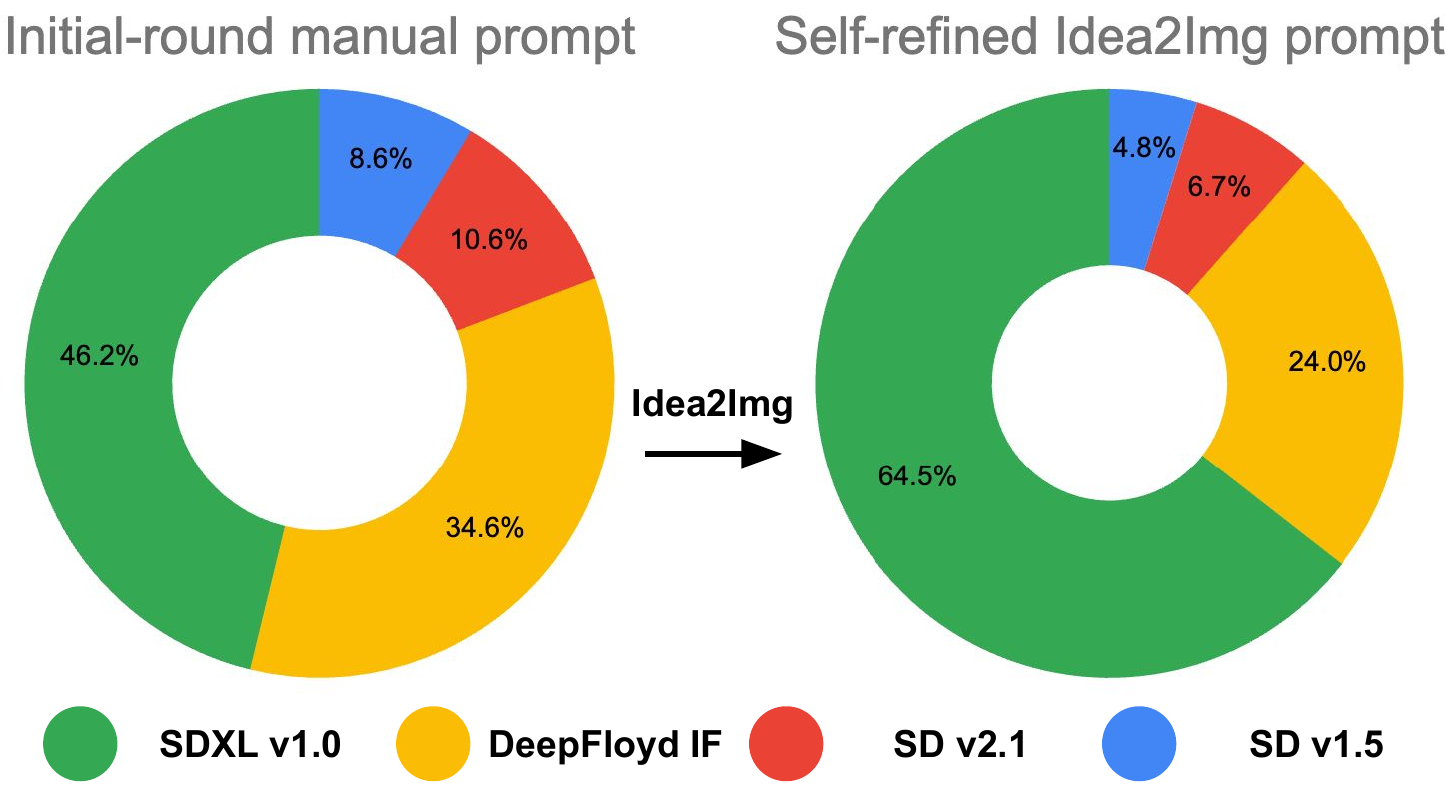}
\caption[Caption for LOF]{
    User preference scores among T2I models before and after iterative self-refinement. We observe that the initially favored T2I model, SDXL, benefits more from the \modelname iteration.
	}
\label{fig:modelscore}
\end{figure}

\noindent\textbf{Experimented T2I models.}
We experiment \modelname on a wide variety of T2I model $\mathcal{G}$ with diverse model capacities and functionalities. Specifically, we study Stable Diffusion (SD) v1.5~\cite{rombach2022high}, SD v2.1, SDXL v1.0 with refiner~\cite{podell2023sdxl}, and DeepFloyd IF (IF-I-XL and IF-II-L)~\cite{Deepfloyd}. Other than T2I models, we also consider the img2img pipeline (\ie, SDEdit~\cite{meng2021sdedit}) for SDXL and DeepFloyd IF, as a demonstration of using \modelname for the text-conditioned image-to-image generation. The default strength $t_0$ in the img2img pipeline is $1.00$. SDXL-img2img and IF-img2img are the same as SDXL and IF (\ie, T2I) when \idea contains text only, and condition on the first image with \idea contains multiple images. LMM prompts $p_{gen}, p_{select}, p_{fb}, p_{revise}$ are kept the same for all experimented T2I models. Appendix Section~\textcolor{red}{B} shows the complete LMM prompts.

\subsection{Image Generation Results}
\noindent\textbf{User preference evaluation.}
Table~\ref{table:user} compares the user preference when selecting from the three images generated by ``initial-round manual prompt,'' ``initial-round \modelname prompt,'' and ``iterative self-refined \modelname prompt,'' for each user \idea with the same T2I model. Among T2I models with different model sizes and functionalities, \modelname leads to consistent improvements in user preference. The initial-round \modelname prompt already improves the initial-round manual prompt, by effectively understanding the multimodal user \idea and selecting the best draft images. The full \modelname framework further improves from the initial-round \modelname results with the multimodal iterative self-refinement. For example, when using SDXL v1.0, users prefer the images generated by \modelname $59/104=56.7\%$ times, compared with the baseline of $14/104=13.5\%$. Similar improvements are observed on all experimented T2I models, as shown in the bolded column ``iterative self-refined \modelname prompt.''

Furthermore, we examine which T2I model benefits the most from the LMM iterative self-refinement. By comparing the $\Delta_\text{iteration}$ in Table~\ref{table:user} that represents the difference between first-round and iterative \modelname user preference, we observe that stronger T2I models tend to benefit more from LMM refinements. For example, SDXL and IF become more favored $26.9\%$ and $16.3\%$ times after iteration, compared with the $5.8\%$ and $4.8\%$ for SD v2.1 and SD v1.5. The trend that stronger T2I models benefit more from \modelname is also observed in Figure~\ref{fig:modelscore}'s analysis, where users pick their preferred image generated by different T2I models. After \modelname's iterative refinement, the initially favored model SDXL benefits more from the iteration, resulting in an even higher user preference rate, from $46.2\%$ to $65.4\%$. We conjecture that the better language understanding ability in stronger T2I models enables them to better follow revised T2I prompts. They also have a better image generation capability that makes it possible to generate challenging scenes, when given a good T2I prompt optimized by \modelname. Nonetheless, \modelname is effective across T2I models of varying capacities, consistently leading to a higher user preference score.

\begin{figure*}[t]
\centering
\includegraphics[width=0.9\textwidth]{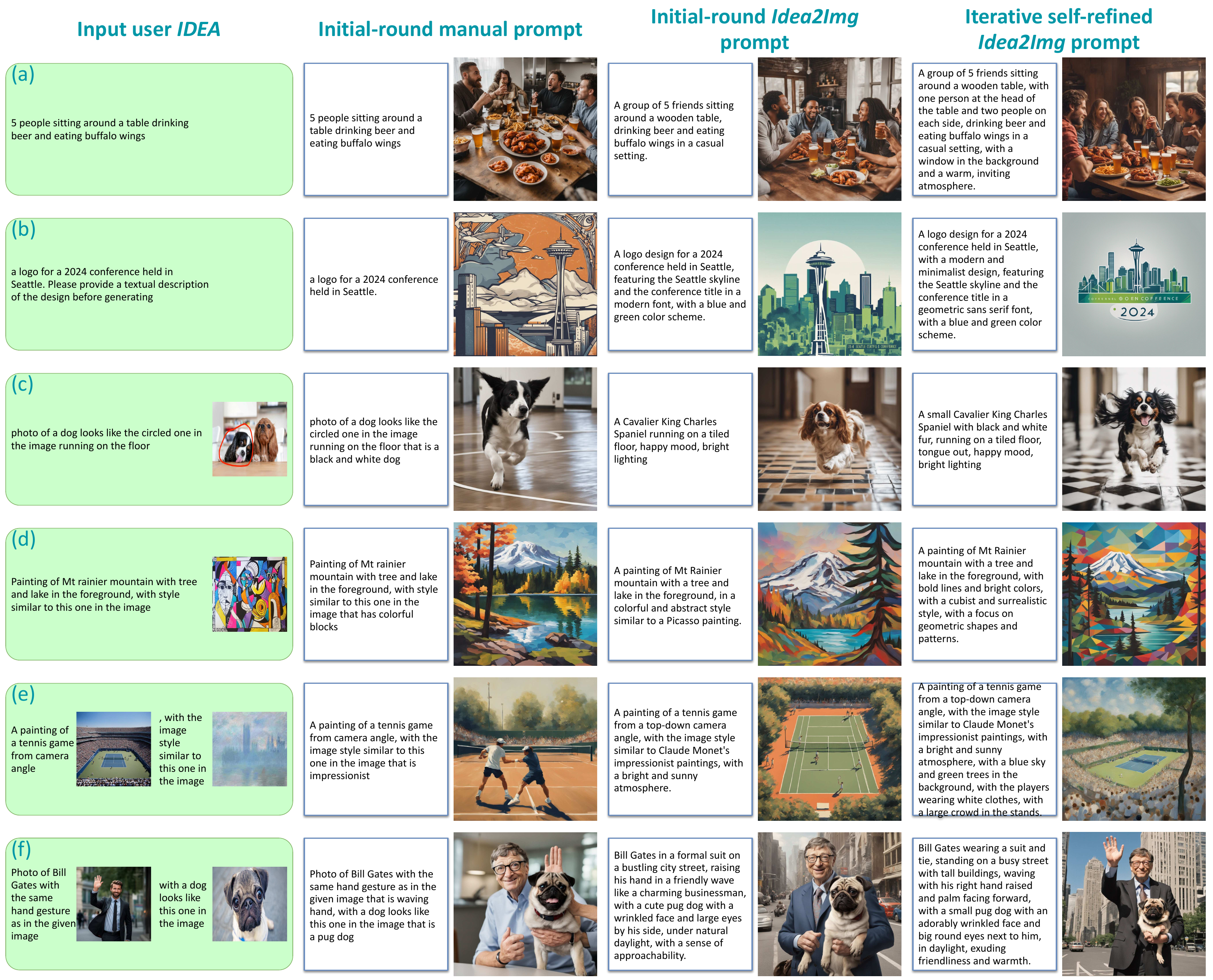}
\caption[Caption for LOF]{
    The comparisons among initial-round manual prompt, initial-round \modelname prompt, and iterative self-refined \modelname prompt, with SDXL~\cite{podell2023sdxl} as the T2I model.
	}
\label{fig:main_results}
\end{figure*}

\begin{figure*}[t]
\centering
\includegraphics[width=0.9\textwidth]{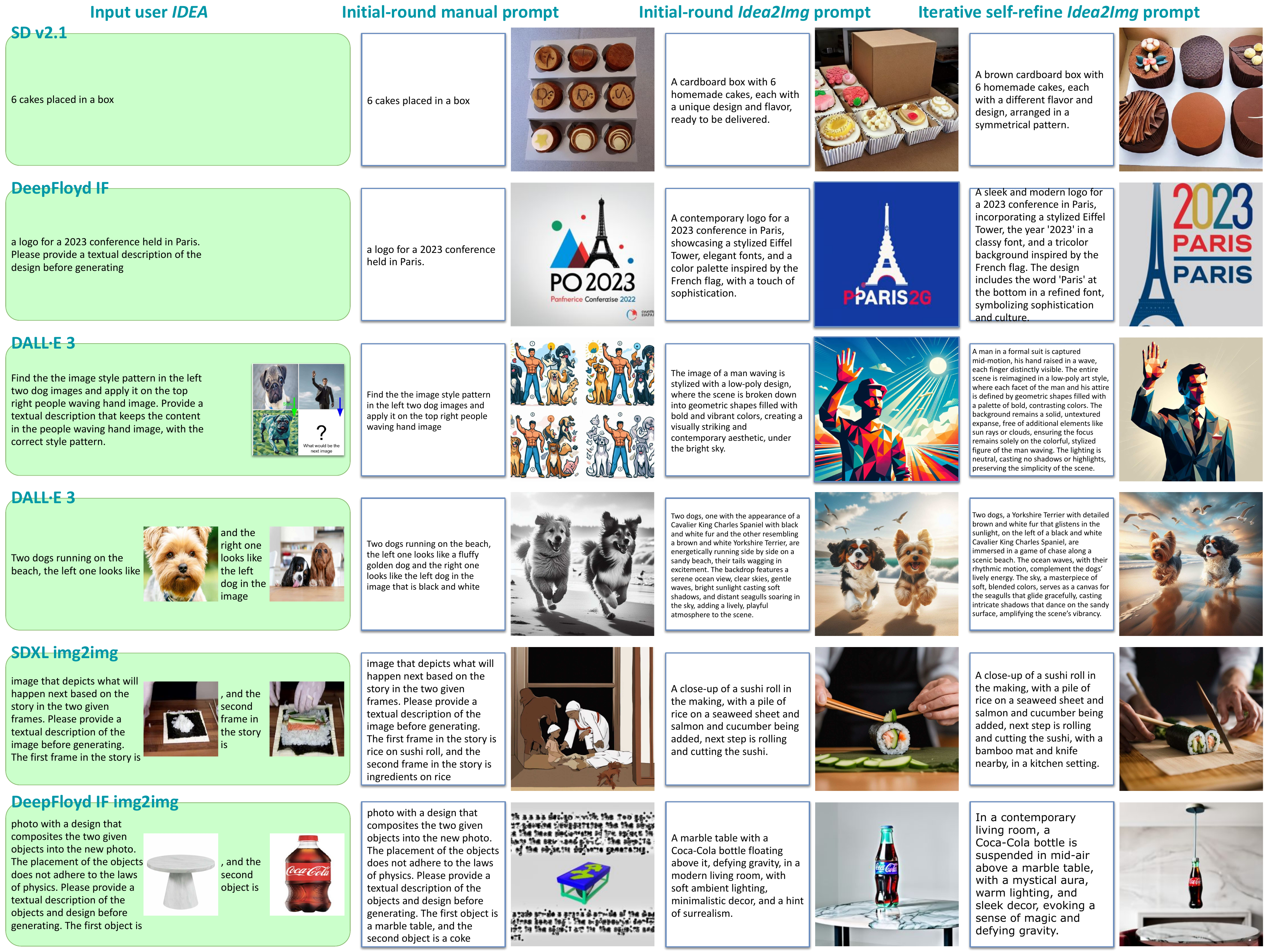}
\caption[Caption for LOF]{
    The comparisons among initial-round manual prompt, initial-round \modelname prompt, and iterative self-refined \modelname prompt, with different image generation models. Additional qualitative results and discussions are in Appendix~\textcolor{red}{A.1}. %
	}
\label{fig:main_results_others}
\end{figure*}

\noindent\textbf{Qualitative comparisons.}
\modelname could help users generate images that better follow \idea, such as the correct object counts in Figure~\ref{fig:main_results}(a). \modelname enables visual content design, in contrast to conventional T2I that requires a detailed visual content description. For example in Figure~\ref{fig:main_results}(b), \modelname designs visual logo based on the instruction of ``a logo for a 2024 conference in Seattle.'' The power of LMMs allows \modelname to extract arbitrary information from the input image for visual generation. This could be any object in the image like ``the circled dog'' in Figure~\ref{fig:main_results}(c) or the image style like in Figure~\ref{fig:main_results}(d). Such general visual conditioning ability can be seamlessly extended to compose multiple visual and text conditions, such as composing the camera angle and image style in Figure~\ref{fig:main_results}(e) and two objects in Figure~\ref{fig:main_results}(f).

Other than SDXL, \modelname is effective in finding text prompts for other image generation models. This includes arbitrary T2I models (\eg, SD v2.1~\cite{rombach2022high}, DeepFloyd IF~\cite{Deepfloyd}, DALL·E 3~\cite{dallecard}, \etc), text-conditioned image-to-image models (\eg, SDXL-img2img and IF-img2img with SDEdit~\cite{meng2021sdedit}), and other specialist generation models (\eg, reward-tuned T2I~\cite{lee2023aligning,fan2024reinforcement}, region-controlled generators~\cite{zhang2023adding,yang2023reco,li2023gligen}, and other specialist models~\cite{avrahami2022blended,ruiz2023dreambooth,brooks2023instructpix2pix}). Figure~\ref{fig:main_results_others} overviews \modelname working with different image generation models. 
We show additional qualitative results and discussions in Appendix Section~\textcolor{red}{A.1}.%

\noindent\textbf{How \modelname may assist humans?}
We use selected qualitative results to highlight the scenarios where humans might find \modelname most helpful in image design and generation, compared with conventional T2I generation.
\begin{enumerate}

\item \textbf{New functionalities with multimodal \idea inputs.} \modelname provides a more natural way for human interaction, where users do not have to describe their desired image solely through texts and painstakingly search for the right prompt word. Instead, the multimodal \idea allows \modelname to precisely extract specific elements from one or multiple input images, such as the dog breed and color, pointed objects, artist style, camera view, and more, as shown in Figure~\ref{fig:main_results}. Finding proper words that the T2I model can understand for such visual concepts could be tedious for humans, \eg, the art style ``with bold lines and bright colors, with a cubist and surrealistic style, with a focus on geometric shapes and patterns.'' in Figure~\ref{fig:main_results}(d). \modelname automates this process via \modelname iterative self-refinement.

\item \textbf{New functionalities with instructional inputs.} 
Vanilla T2I models struggle to understand T2I prompts that describe the intended visual design or purpose of the generated image, such as ``a logo for a 2024 conference held in Seattle'' in Figure~\ref{fig:main_results}(b). Instead, the prompt needs to be a comprehensive description of the image to generate, demanding extra drafting effort from users, such as ``\ldots the Seattle skyline in the center and the conference title below it \ldots''. In contrast, \modelname effectively understands the instructional texts in \idea and creates images accordingly.

\item \textbf{Better semantic and visual quality.} Finally, the iterative refinement allows \modelname to generate images with better semantic and visual qualities, leading to an effective automatic image creation assistant.
\end{enumerate}

\begin{figure*}[t]
\captionsetup[subfloat]{}
\centering
\subfloat[\textbf{Feedback reflection:} The right column shows the examples of generated text feedback.]{\includegraphics[width=0.93\textwidth]{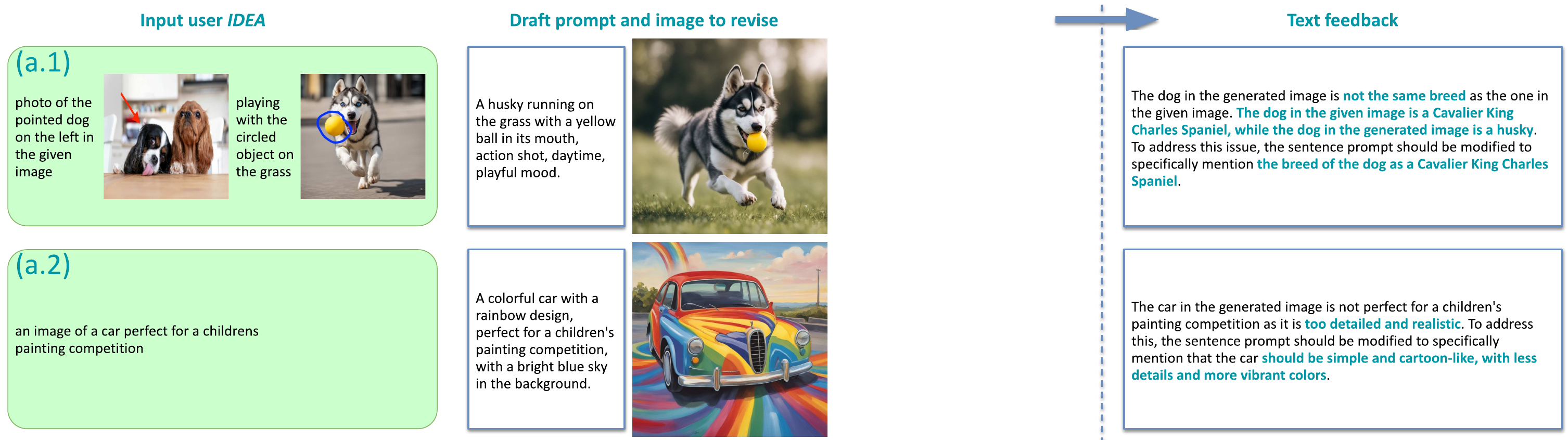}}%

\subfloat[\textbf{Revised prompt generation:} The right column shows the examples of revised prompts.]{\includegraphics[width=0.93\textwidth]{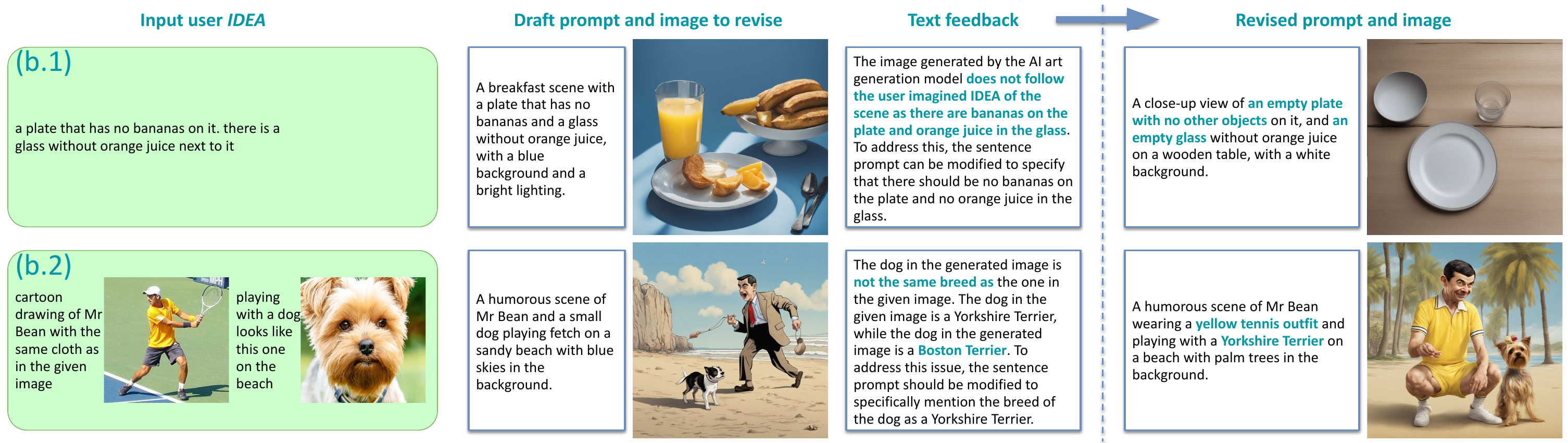}}%

\subfloat[\textbf{Draft image selection:} The right column shows the examples of the draft image selection index and justification.]{\includegraphics[width=0.93\textwidth]{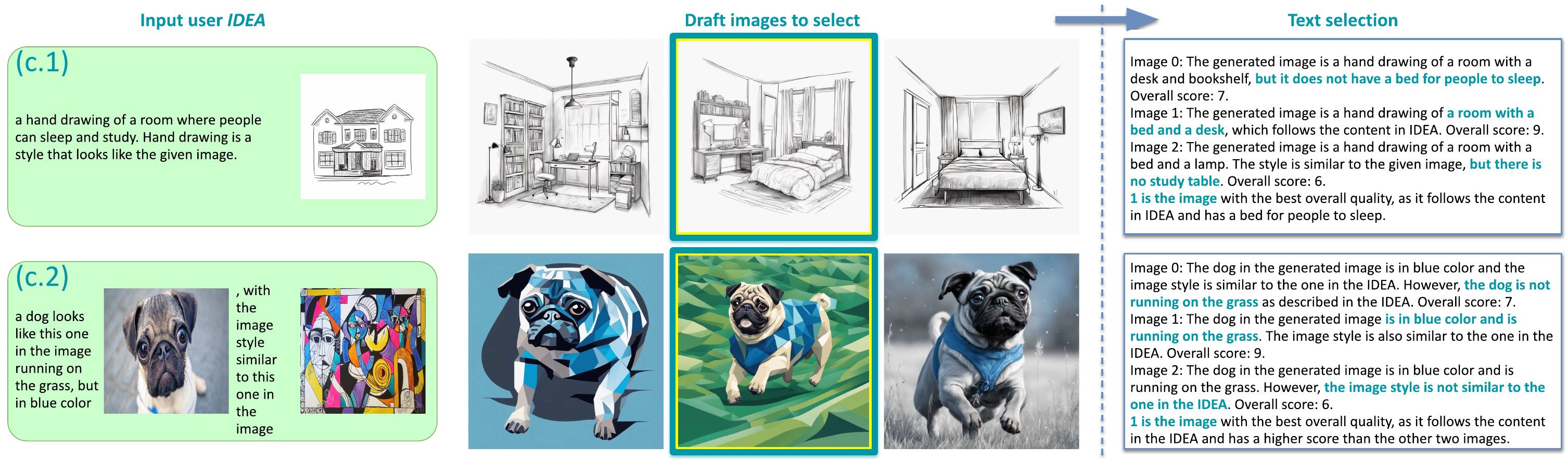}}%
\caption[Caption for LOF]{\lmmname's outputs in \modelname for image feedback, revision, and selection.}
\label{fig:gpt4_lmmoutput}
\end{figure*}

\subsection{LMM Feedback, Revision, and Selection}
We show representative LMM outputs for ``feedback reflection,'' ``revised prompt generation,'' and ``draft image selection.'' Additional results are in Appendix~\textcolor{red}{A.2}.%

\noindent\textbf{Feedback reflection.}
Figure~\ref{fig:gpt4_lmmoutput}(a) shows the text feedback generated by \lmmname for the user \idea and the draft image and T2I prompt. \modelname can effectively check if the generated image is correct, and verify if the draft image corresponds to the visual descriptions in \idea. This includes the breed of the dog in (a.1), as well as art styles, objects, visual attributes, \etc. In addition to identifying the discrepancy, \modelname also points to the plausible directions that may improve the T2I prompt in the text feedback. For example, in (a.2), \modelname provides guidance to have generated images better follow the user intention of ``an image for a children's painting competition,'' by ``specifically mentioning that the car should be simple and cartoon-like.''

\noindent\textbf{Revised prompt generation.}
Figure~\ref{fig:gpt4_lmmoutput}(b) compares the T2I prompts before and after the prompt revision, showcasing how text feedback may help the refinement. For example, in (b.1), the revised T2I prompt specifies ``an empty plate with no other objects'' to preclude the T2I model from generating bananas, which occurred with the previous prompt ``no bananas.'' In (b.2), the revised T2I prompt includes a detailed description of ``a yellow tennis outfit'' and ``a Yorkshire Terrier'' to generate the queried clothing and dog. %

\noindent\textbf{Draft image selection.}
Performing draft image selection requires LMMs to compare multiple similar draft images and pick the one that best matches the multimodal input \idea. Figure~\ref{fig:gpt4_lmmoutput}(c) shows the selection results generated by \modelname. \lmmname is prompted to give justifications and scores for each draft image, in addition to the final selection. 
We observe that \modelname could comprehensively judges different aspects in \idea, and gives reasonable scores and selection indexes. \Eg, finding the image with both sleep and study area in (c.1), verifying content and style in (c.2), and other examples in Appendix Figure~\textcolor{red}{G}.%

\section{Limitation and Discussion}
\label{appd:limit}
\noindent\textbf{Tasks beyond image generation.}
\modelname explores the emergent ability of multimodal self-refinement in LMM-based systems, through the image design and generation task. Specifically, \modelname views the T2I model to use as an unknown multimodal environment to explore, and iteratively refines T2I prompts to find its optimal usage. This concept mirrors the intrinsic human approach of iterative problem-solving when faced with unknown environments or challenges. We leave its extension to other intriguing tasks, \eg, GUI navigation~\cite{yan2023gpt}, embodied agents~\cite{nasiriany2024pivot}, and complicated visual reasoning~\cite{wu2023v,qi2024cogcom}, for future exploration.

\noindent\textbf{From a single image generation model to multiple tools.}
\modelname explores using a single image generation model, such as a text-to-image model~\cite{rombach2022high} or a text-conditioned image-to-image model~\cite{meng2021sdedit}. When needed, other specialized generative models like ControlNet~\cite{zhang2023adding}, inpainting~\cite{avrahami2022blended}, region-controlled T2I generation~\cite{yang2023reco,li2023gligen}, customized generation~\cite{ruiz2023dreambooth,chen2023subject}, and video generation~\cite{singer2022make,yin2023nuwa} can be seamlessly switched and supported. That is, \modelname could broadly boost different visual generation models of diverse specialties by exploring their optimal text description or instruction prompts. 
Beyond a single generation model, \modelname can also be used to allocate multiple tools as in multimodal agent studies~\cite{yang2023mmreact,wu2023visual}. In this case, \modelname isn't limited to optimizing the use of individual tools but also investigates their effective collaboration when used together, such as generator selection and multi-step visual generation. %

\noindent\textbf{Consolidating explored knowledge.}
We have shown the effectiveness of LMM iterative self-refinement in automatic image design and generation. 
\modelname can also help to consolidate or distill the explored knowledge into T2I model parameters, such that no inference-time iterative refinement is needed when encountering seen generation scenarios. One could collect a dataset using \modelname for a scenario of interest, and fine-tune a T2I model with the explored self-refinement trajectory. Storing the probed knowledge as sample-agnostic prompt for each image generation model is another promising direction~\cite{zhao2023expel,pryzant2023automatic,guo2023learning}. Finally, with minimal extra computation, we find it helpful to use the explored T2I prompt history as in-context examples for prompt re-writing and expansion, improving from the zero-shot expansion like the one in ChatGPT-Dalle-3~\cite{betker2023improving,gpt4vblog}.

\section{Conclusion}
We have presented \modelname, a multimodal iterative self-refinement framework that leverages \lmmnamefull for image design and generation. \modelname explores the emergent capabilities of iterative self-refinement in LMM-based agent systems, showcasing its effectiveness in improving, assessing, and verifying the generated multimodal content.  The user preference study demonstrates \modelname's capability in assisting humans to find the optimal usage of generation models for automatic image design and generation.

\subsection*{Acknowledgment}

We are deeply grateful to OpenAI for providing access to their exceptional tool~\cite{openai2023gpt4,gpt4v,gpt4vcontribution,gpt4vblog}. We also extend heartfelt thanks to our Microsoft colleagues for their insights, with special acknowledgment to Faisal Ahmed, Ehsan Azarnasab, and Lin Liang for their constructive feedback.

\appendix
\setcounter{figure}{0} %
\renewcommand{\thefigure}{\Alph{figure}} %
\setcounter{table}{0} %
\renewcommand{\thetable}{\Alph{table}} %

\begin{figure*}[t!]
\centering
\vspace{0.2in}
\centerline{\includegraphics[width=1.1\textwidth]{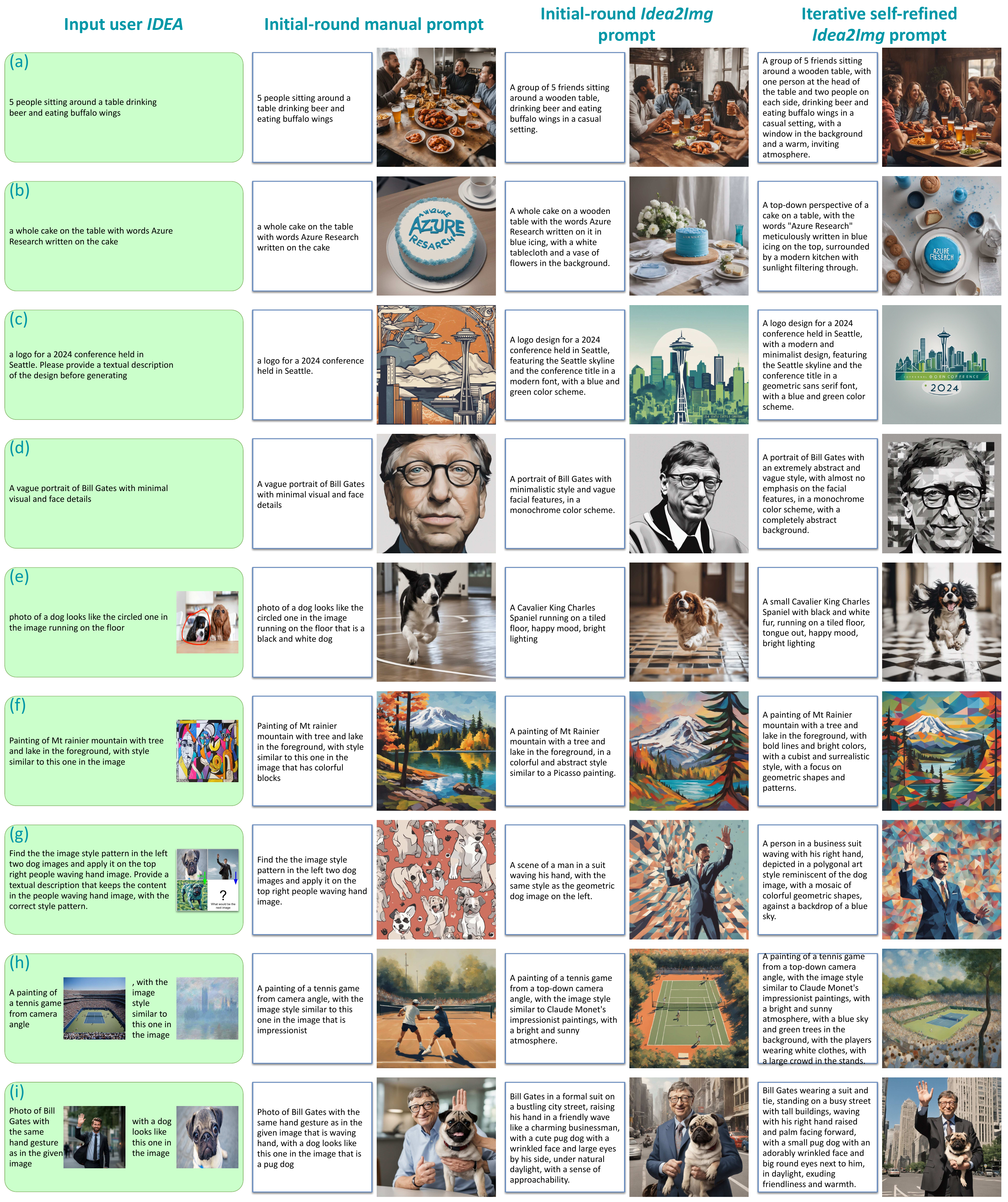}}
\caption[Caption for LOF]{
    The comparisons among the initial-round manual prompts, initial-round \modelname prompts, and the iterative self-refined \modelname prompts, with the SDXL v1.0~\cite{podell2023sdxl} used as the T2I model.
	}
\label{fig:main_sdxl}
\end{figure*}

\begin{figure*}[t!]
\centering
\vspace{0.2in}
\centerline{\includegraphics[width=1.1\textwidth]{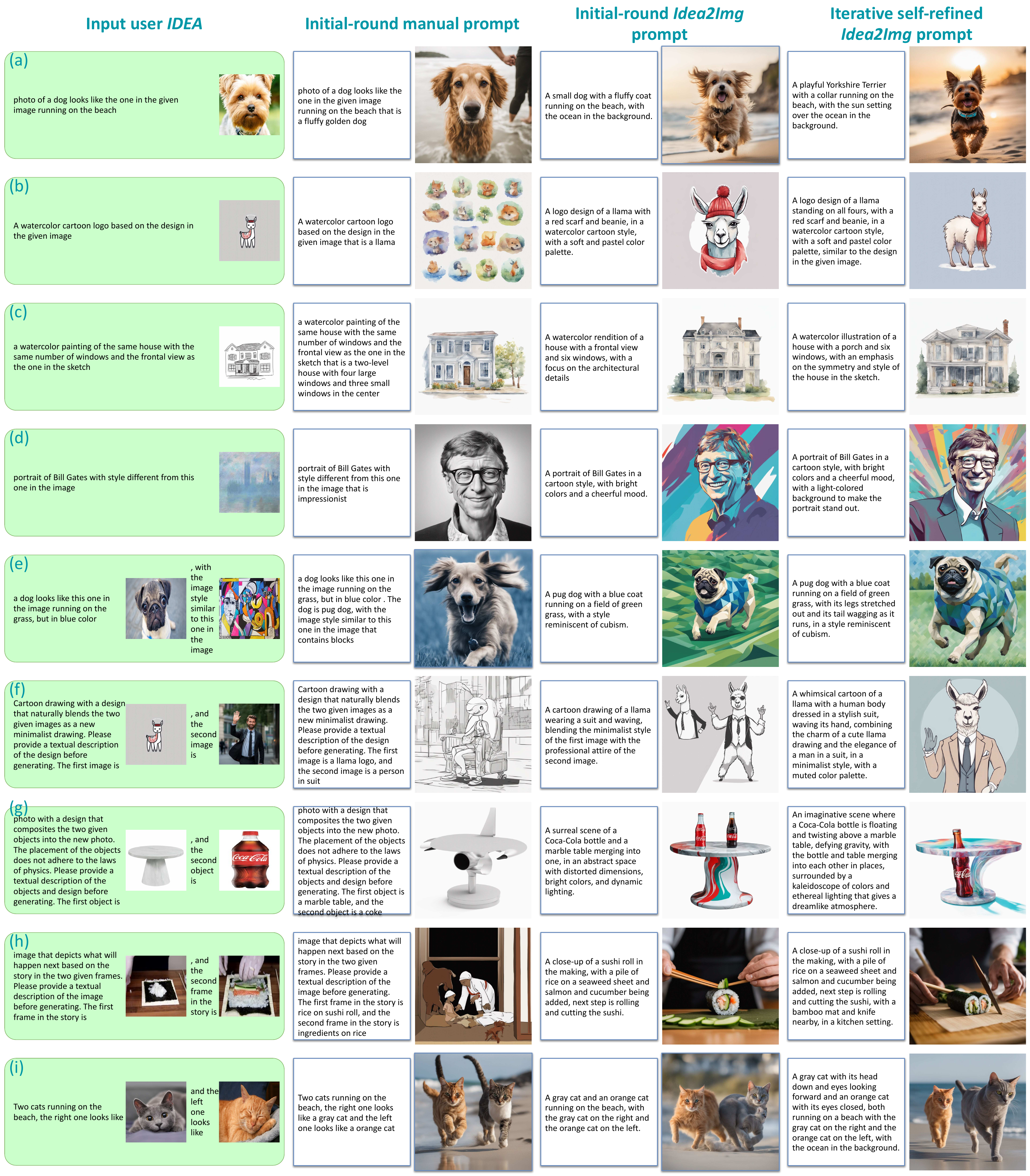}}
\caption[Caption for LOF]{
    The comparisons among the initial-round manual prompts, initial-round \modelname prompts, and the iterative self-refined \modelname prompts, with the SDXL-img2img~\cite{podell2023sdxl,meng2021sdedit} used as the image generation model. Instead of random noise, the image generation starts from the input image with added noise~\cite{meng2021sdedit}, showing the effectiveness of \modelname on text-conditioned image-to-image pipelines. 
	}
\label{fig:main_sdxlimg2img}
\end{figure*}

\begin{figure*}[t!]
\vspace{0.2in}
\centering
\centerline{\includegraphics[width=1.1\textwidth]{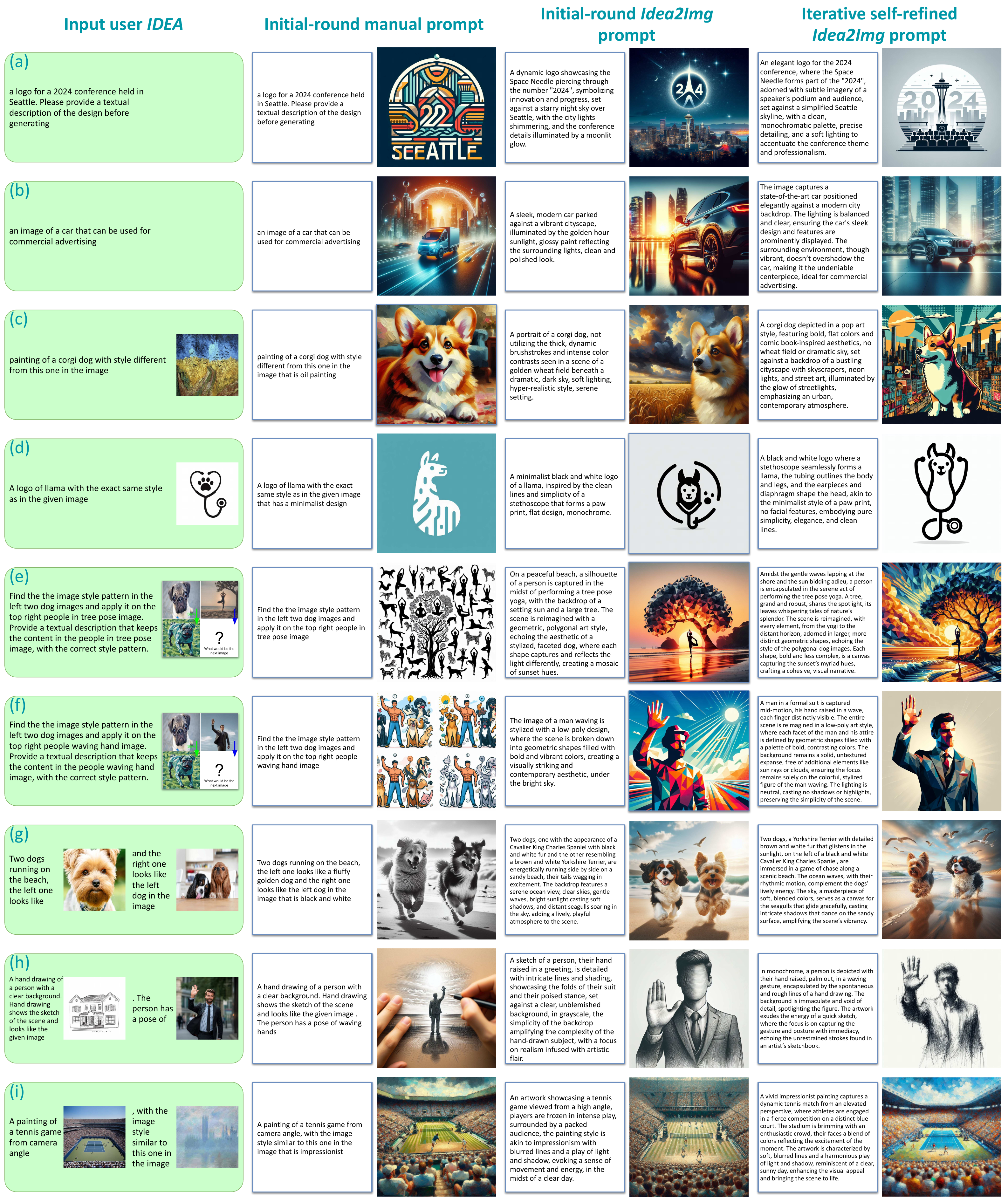}}
\caption[Caption for LOF]{
    The comparisons among the initial-round manual prompts, initial-round \modelname prompts, and the iterative self-refined \modelname prompts, with the Dalle-3~\cite{dallecard} used as the T2I model.
	}
\label{fig:main_de3}
\end{figure*}

\begin{figure*}[t!]
\centering
\centerline{\includegraphics[width=1.1\textwidth]{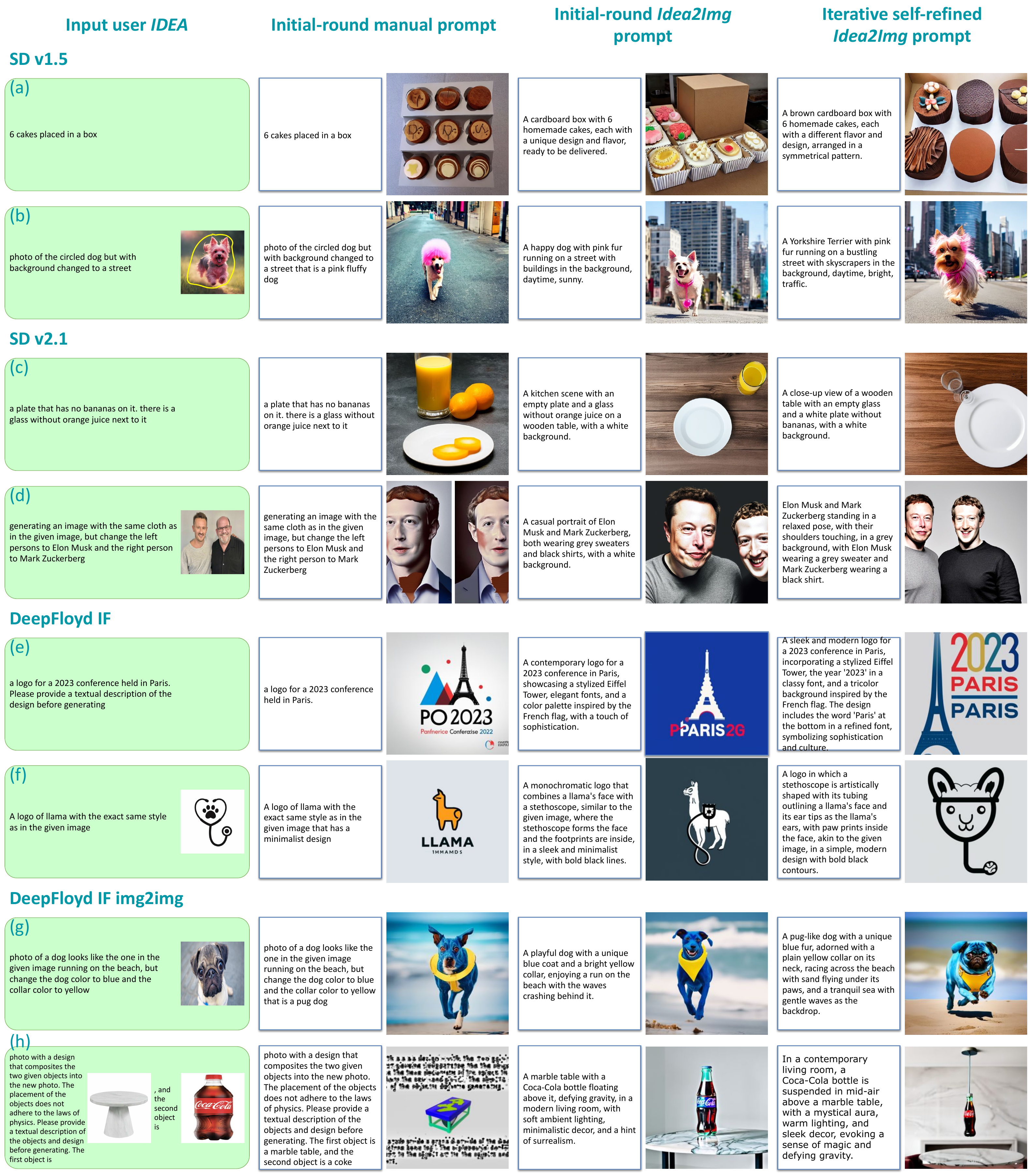}}
\caption[Caption for LOF]{
    The comparisons among the initial-round manual prompts, initial-round \modelname prompts, and the iterative self-refined \modelname prompts, with other image generation models, including SD v1.5, SD v2.1~\cite{rombach2022high}, DeepFloyd IF, and IF-img2img~\cite{Deepfloyd}.
	}
\label{fig:main_others}
\end{figure*}
In this supplementary material, we begin with showing additional qualitative results in Section~\ref{appd:example}, in supporting \modelname's effectiveness on different image generation models, including Dalle-3~\cite{betker2023improving,dallecard}, SDXL~\cite{podell2023sdxl}, SDXL-img2img~\cite{podell2023sdxl,meng2021sdedit}, DeepFloyd IF~\cite{Deepfloyd}, among others. In Section~\ref{appd:lmm_output}, we show \lmmname's outputs to probe how \modelname helps image creation during the iterative self-refinement, and the possibility of replacing \lmmname with other LMMs. %
Section~\ref{appd:code} introduce remaining implementation details.

\section{Qualitative Results}
\subsection{Qualitative Comparisons}
\label{appd:example}
Figures~\ref{fig:main_sdxl}-\ref{fig:main_others} show additional qualitative results of the comparison in Table~\textcolor{red}{1}. %
Figure~\ref{fig:main_sdxl} presents examples of \modelname explores the use of SDXL, a representative T2I model. Figure~\ref{fig:main_sdxlimg2img} examines SDXL-img2img, a simple text-conditioned image-to-image model that adds noise to the input image and then performs text-conditioned denoising~\cite{meng2021sdedit}. Figures~\ref{fig:main_de3},~\ref{fig:main_others} contain the results of \modelname working with Dalle-3 and other image generation models.

\vspace{1mm}
\noindent\textbf{SDXL.}
\modelname could help users generate images that better follow \idea, such as the one with correct object counts and rendered scene texts in Figures~\ref{fig:main_sdxl}(a,b). \modelname enables the visual content design that can create images from a text instruction of its desired usage, in contrast to the detailed image description required in the conventional T2I generation.
For example in Figure~\ref{fig:main_sdxl}(c), \modelname designs a logo based on the user \idea of ``having a logo for a 2024 conference in Seattle.'' \modelname can also understand user \idea to search for images with high aesthetic scores and great visual details, or its opposite direction with ``minimal face details'' in (d). The LMM allows \modelname to extract arbitrary information from the input image for visual generation. This could be any specific object in the image, such as ``the dog on the left'' or ``the dog pointed to via a red circle'' in (e). Figure~\ref{fig:main_sdxl}(f) shows an example of extracting the painting style, which requires art knowledge for humans to describe accurately. The image input can even be an in-context example that defines the desired image transformation, such as the visual style transfer shown in (g). The ability to extract arbitrary information from the input image can be seamlessly extended to compose multiple visual and text conditions, such as composing the camera angle and image style in (h) and the two entities in (I).

\vspace{1mm}
\noindent\textbf{SDXL-img2img.}
\modelname is also effective in finding T2I prompts for the text-conditioned image-to-image model SDXL-img2img, as shown in Figure~\ref{fig:main_sdxlimg2img}. Figures~\ref{fig:main_sdxlimg2img}(c) and (d) illustrate generating images that follow and differ from the reference image style of ``watercolor and impressionist,'' respectively. \modelname can design visual contents with the inspiration of input images, \eg, a cartoon drawing that blends the llama and the person in suits in (f), and composing the coke with the table in an imaginative way in (g). (h) illustrates a novel scenario of generating an image to represent the anticipated action of rolling sushi.

\vspace{1mm}
\noindent\textbf{Dalle-3 and other generation models.}
\modelname demonstrates its effectiveness across different image generation models. Figure~\ref{fig:main_de3} shows the results generated by \modelname with Dalle-3. We access Dalle-3 via Bing Image Creator\footnote{\url{https://www.bing.com/create}}, which excludes the ChatGPT prompt rewrite. \modelname could better release Dalle-3's strong prompt-following capability and show impressive results, especially for challenging queries. This includes polishing the logo design in Figure~\ref{fig:main_de3}(a), drafting car advertisements in (b), creating unique image styles in (c), and enhancing the design with reference images in (d). When confronted with more challenging tasks, \modelname with Dalle-3 excels. For the visual in-context generation problem in (e) and (f), \modelname finds the pattern in the input grid image and explores T2I prompts for the desired image design. The framework also proves effectiveness when handling multiple reference images, such as the two dogs in (g), the hand drawing of a person's pose in (h), and the tennis game with a queried style in (i).

Furthermore, Figure~\ref{fig:main_others} shows the \modelname results on other T2I models, including SD v1.5, v2.1, DeepFloyd IF, and IF-img2img. Despite the variance in the base T2I models' capacity, \modelname consistently helps design and generate better images.

\begin{figure*}[t!]
\centering
\includegraphics[width=1.05\textwidth]{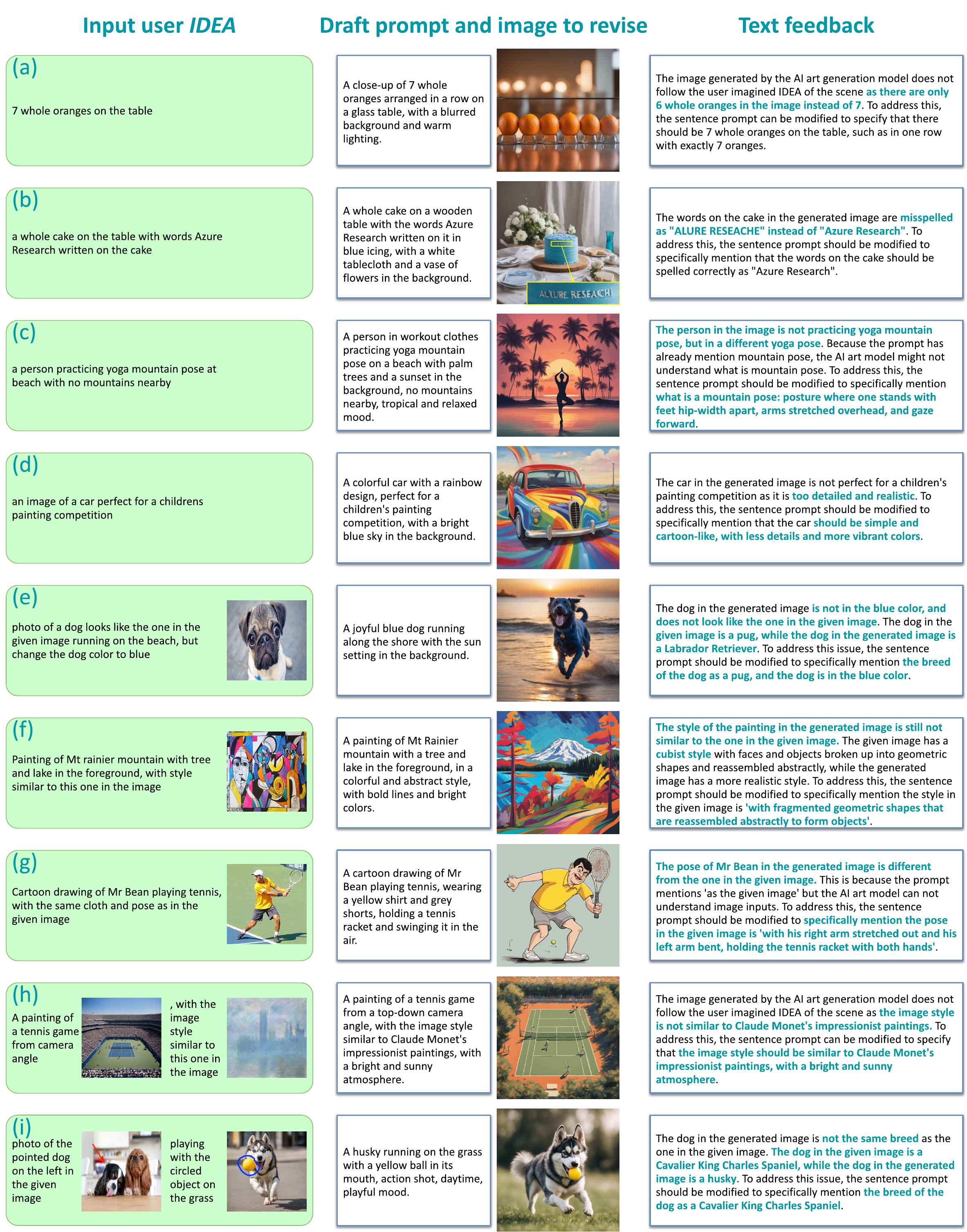}
\caption[Caption for LOF]{
    Examples of the generated text feedback. The left column shows the multimodal input user \idea, and the center column shows the draft image to process as well as its corresponding text prompts. The right column shows the text feedback generated by \lmmname. The {\color{darkblue} 
    \bf dark blue color} highlights the identified discrepancies.
	}
\label{fig:gpt4_prompt}
\end{figure*}

\begin{figure*}[t!]
\centering
 \vspace{0.3in}
\includegraphics[width=1.05\textwidth]{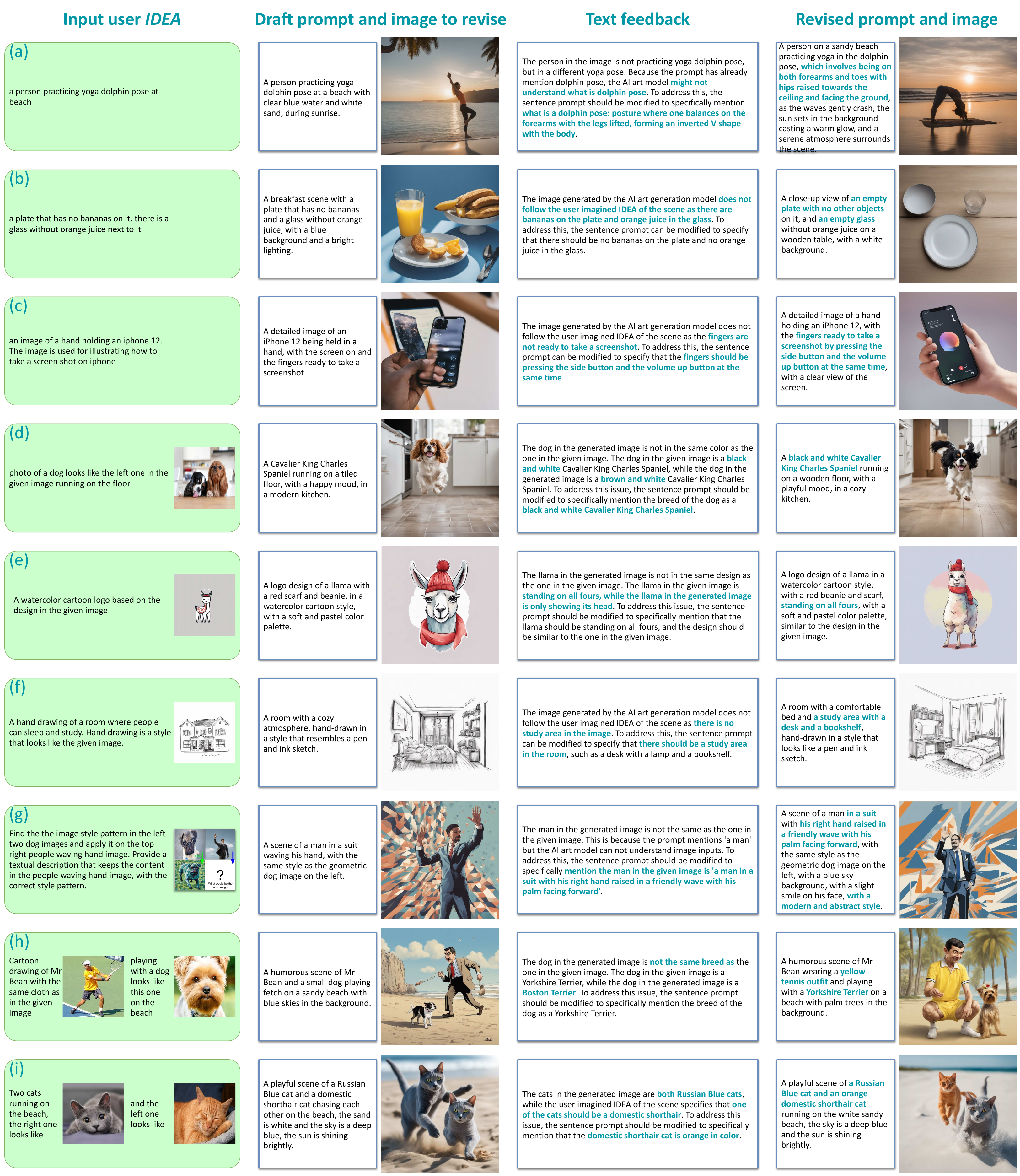}
\caption[Caption for LOF]{
    Examples of the revised prompts. The four columns, from left to right, show the input user \idea, the draft image to be revised, generated text feedback, and the revised T2I prompt and image. The {\color{darkblue} 
    \bf dark blue color} highlights the identified discrepancies in text feedback, and how they are addressed in the revised T2I prompt.
    We note that the example only shows a single round of self-refinement. Therefore, the revised T2I prompt may have remaining issues to be further addressed.
	}
 \vspace{0.1in}
\label{fig:gpt4_revise}
\end{figure*}

\begin{figure*}[t!]
\centering
 \vspace{0.3in}
\includegraphics[width=1.05\textwidth]{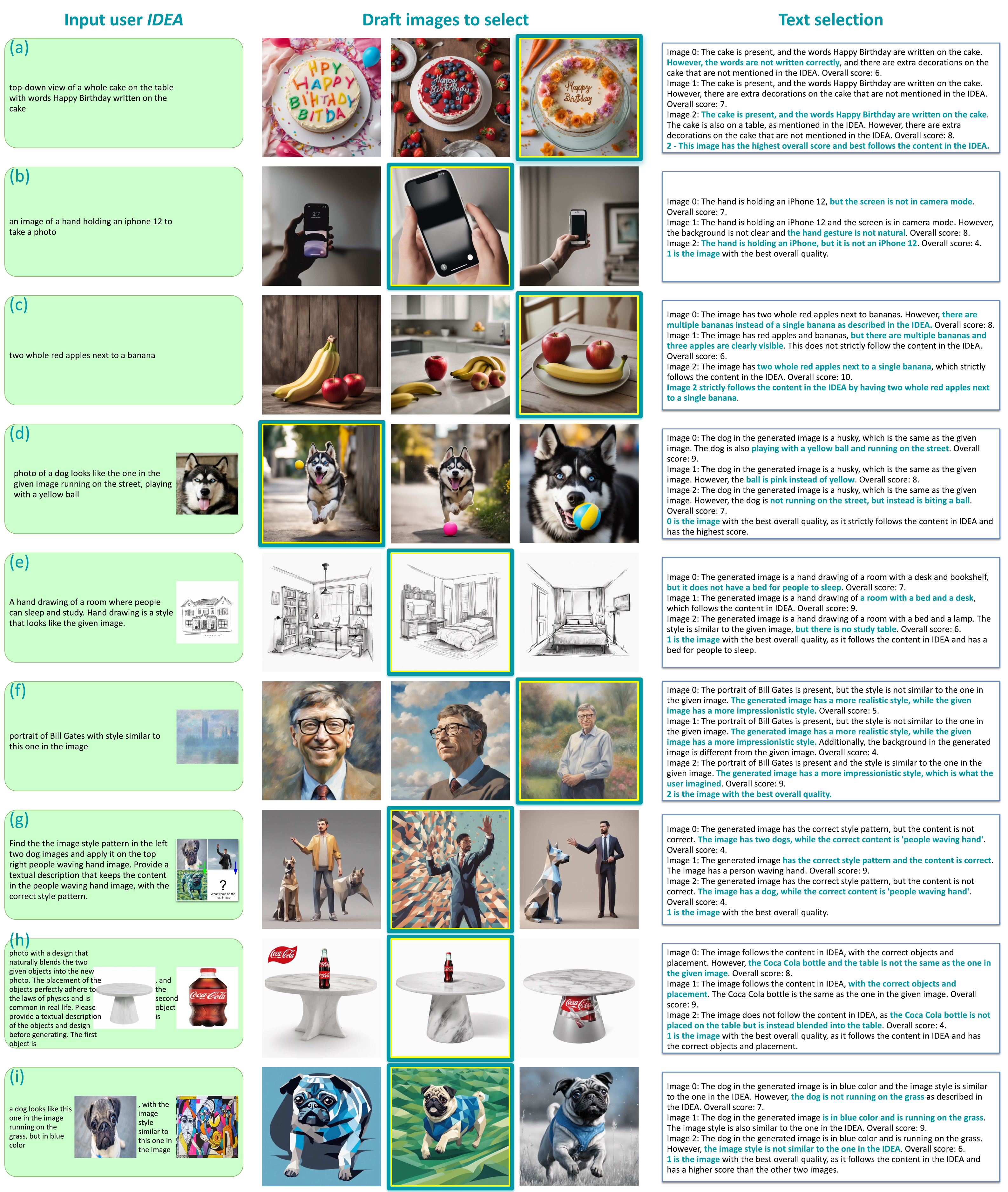}
\caption[Caption for LOF]{
    Examples of the draft image selection. The right column shows justification ratings and the final selection. The {\color{darkblue} 
    \bf dark blue color} highlights the identified problems and justifications for the draft image selection. The selected draft image in the round is visualized with the {\color{darkblue} 
    \bf blue box} shown in the middle column. %
	}
 \vspace{0.1in}
\label{fig:gpt4_select}
\end{figure*}

\begin{figure*}[t]
\captionsetup[subfloat]{}
\centering
\subfloat[\textbf{Feedback reflection:} The right column shows the examples of the generated text feedback.]{\includegraphics[width=\textwidth]{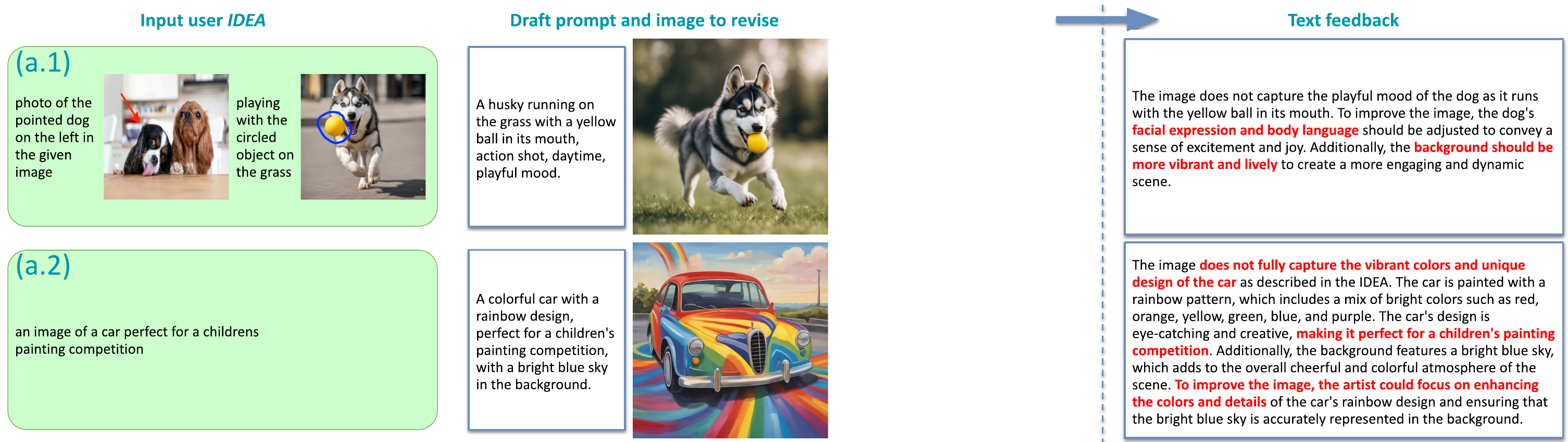}}

\subfloat[\textbf{Revised prompt generation:} The right column shows the examples of the revised prompts.]{\includegraphics[width=\textwidth]{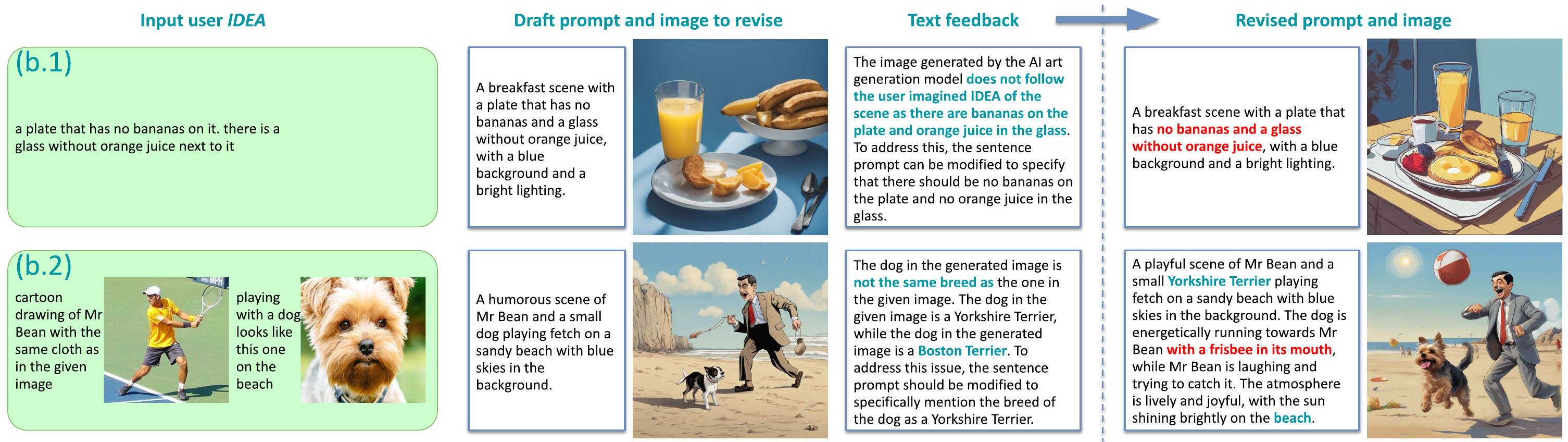}}
\caption[Caption for LOF]{LLaVA-1.5-13B's~\cite{liu2023improved} outputs in \modelname for image feedback and revision.}
\label{fig:gpt4_llavaoutput}
\end{figure*}

\subsection{LMM Feedback, Revision, and Selection}
\label{appd:lmm_output}
One may wonder how \lmmname behaves and performs in each role throughout 
\modelname's iterative self-refinement pipeline, \ie, ``feedback reflection,'' ``revised prompt generation,'' and ``draft image selection.'' We show corresponding qualitative results as follows.

\vspace{1mm}
\noindent\textbf{Feedback reflection.}
Figure~\ref{fig:gpt4_prompt} shows text feedback generated by \lmmname for the user \idea, draft image, and T2I prompt. \modelname can effectively check if the generated image is correct, such as the number of oranges in (a) and the misspelled scene text "ALURE RESEACHE" in (b). In addition to the text descriptions in \idea, \modelname can verify if the draft image corresponds to the visual descriptions in \idea. This includes the color and breed of the dog in (e), the exact art style in (f), and the same cloth and pose in (g). Furthermore, \modelname can understand and verify the interleaved image-text pairs in \idea, as shown in Figures~\ref{fig:gpt4_prompt}(h,i).

In addition to identifying the discrepancy, \modelname can also point to the plausible directions for improving the T2I prompt in the text feedback. For example, in Figure~\ref{fig:gpt4_prompt}(c), \lmmname mentions that ``the person is not in yoga mountain pose, but the T2I prompt has already mentioned mountain pose,'' ``the AI model might not understand what mountain pose is, and prompt should be modified to specifically mention what mountain pose is.'' Similarly, in Figure~\ref{fig:gpt4_prompt}(d), \modelname provides guidance to have generated images better follow the user intention of ``an image for a children's painting competition,'' by ``specifically mentioning that the car should be simple and cartoon-like.''

\vspace{1mm}
\noindent\textbf{Revised prompt generation.}
Figure~\ref{fig:gpt4_revise} compares the T2I prompts before and after the revision, for visualizing how text feedback helps the revision. For example, (a) the revised T2I prompt includes a detailed description of the ``yoga dolphin pose'' to generate the correct body pose; (b) the revised T2I prompt mentions ``an empty plate with no other objects'' to avoid the T2I model misunderstand the prompt ``no bananas;'' (c) T2I model generates the correct hand gesture with \modelname providing text description on how to take a screenshot. \modelname also effectively addresses the identified errors in text feedback and improves the prompts for multimodal input \idea, including the dog color in Figure~\ref{fig:gpt4_revise}(d), the llama design in Figure~\ref{fig:gpt4_revise}(e), the study area in Figure~\ref{fig:gpt4_revise}(f), the human gesture in Figure~\ref{fig:gpt4_revise}(g), the dog breed and human clothing in Figure~\ref{fig:gpt4_revise}(h), and the color of the two cats in Figure~\ref{fig:gpt4_revise}(i).

\vspace{1mm}
\noindent\textbf{Draft image selection.}
T2I models may generate low-quality images even with good T2I prompts. To ensure consistent improvements in each iteration, it is critical to reduce such generation noise by selecting from multiple draft images in each round. Performing such selection requires \lmmname to compare multiple similar draft images and pick the one with the best overall quality. Figure~\ref{fig:gpt4_select} shows the selection results generated by \lmmname. The LMM prompt is designed such that \lmmname gives justifications and scores for each draft image, in addition to the final selection index. Such intermediate thoughts not only help humans interpret the selection process, but also serve as the chain of thought to improve the selection performance.
We observe that \lmmname can compare different aspects mentioned in the \idea and give reasonable scores and selection index. For example, checking the scene text spelling in Figure~\ref{fig:gpt4_select}(a); verifying the phone screen and model in Figure~\ref{fig:gpt4_select}(b); counting the number of apples and bananas in Figure~\ref{fig:gpt4_select}(c); verifying the ball color and dog action in Figure~\ref{fig:gpt4_select}(d); finding the image with both sleep and study area in Figure~\ref{fig:gpt4_select}(e); selecting the image that best fits the given image style in Figure~\ref{fig:gpt4_select}(f); verifying the image content and style in Figure~\ref{fig:gpt4_select}(g); locating the best-blended image in Figure~\ref{fig:gpt4_select}(h); and finding the image with correct dog color and image style in Figure~\ref{fig:gpt4_select}(I).

\vspace{1mm}
\noindent\textbf{LMMs alternative to \lmmname.}
After observing the effectiveness of \modelname with \lmmname, a natural question is whether we can replace \lmmname with more accessible and lightweight alternatives. Figure~\ref{fig:gpt4_llavaoutput} examines 	
LLaVA-1.5-13B~\cite{liu2023improved,liu2023visual}, a leading open-source LMM, using the same test cases as those in the main paper's Figure~\textcolor{red}{6}. Despite the promising results, LMMs alternative to \lmmname may not be ready yet for the \modelname-like iterative self-refinement framework, with two major bottlenecks. First, most current LMMs lack the ability to process complex interleaved image-text sequences, therefore limiting \modelname in understanding multimodal \idea, processing memory, and referencing in-context examples. This limitation also prevents us from conducting image selection experiments in Figure~\ref{fig:gpt4_llavaoutput}, as we did in Figure~\textcolor{red}{6}(c) with \lmmname. Second, the weaker multimodal reasoning capability~\cite{yu2023mm} will significantly increase the noise in \modelname's iteration and make the framework ineffective. For example, in Figure~\ref{fig:gpt4_llavaoutput}(a.2), LLaVA fails to capture the correct direction to improve the image, and in (b.1), it repeats the same T2I prompt without effective revision.

\section{\modelname Code, Data, and Gallery}
\label{appd:code}
We will release the \modelname code, evaluation queries, and generated samples. 

We show the used LMM prompts $p_{gen}, p_{select}, p_{fb}, p_{revise}$ as follows. {\color{darkblue} \textbf{{The colored texts}}} indicate the corresponding multimodal contents, such as \idea or the history memory. LMM prompts are kept the same for different image generation models and input \idea.

\vspace{8pt}
\textbf{Initial prompt generation $p_{gen}$:}

\begin{center}
\begin{myquote}
{\footnotesize
You are a helpful assistant.
\newline

Instruction: Given a user imagined IDEA of the scene, converting the IDEA into a self-contained sentence prompt that will be used to generate an image.

Here are some rules to write good prompts:

- Each prompt should consist of a description of the scene followed by modifiers divided by commas.

- The modifiers should alter the mood, style, lighting, and other aspects of the scene.

- Multiple modifiers can be used to provide more specific details.

- When generating prompts, reduce abstract psychological and emotional descriptions.

- When generating prompts, explain images and unusual entities in IDEA with detailed descriptions of the scene.

- Do not mention 'given image' in output, use detailed texts to describe the image in IDEA instead.

- Generate diverse prompts.

- Each prompt should have no more than 50 words.

IDEA: {\color{darkblue} \textbf{{\idea input}}}.

End of IDEA.

Based on the above information, you will write {\color{darkblue} \textbf{{N}}} detailed prompts exactly about the IDEA follow the rules. Each prompt is wrapped with <START> and <END>.
}
\end{myquote}
\end{center}

\vspace{8pt}
\textbf{Draft image selection $p_{select}$:}

\begin{center}
\begin{myquote}
{\footnotesize
You are a helpful assistant.
\newline

You are a judge to rank provided images. Below are {\color{darkblue} \textbf{{N}}} images generated by an AI art generation model, indexed from 0 to {\color{darkblue} \textbf{{N-1}}}.

From scale 1 to 10, decide how similar each image is to the user imagined IDEA of the scene.

IDEA: {\color{darkblue} \textbf{{\idea input}}}.

End of IDEA.

{\color{darkblue} \textbf{{List of draft images.}}}

Let's think step by step. Check all aspects to see how well these images strictly follow the content in IDEA, including having correct object counts, attributes, entities, relationships, sizes, appearance, and all other descriptions in the IDEA. Then give a score for each input images. Finally, consider the scores and select the image with the best overall quality with image index 0 to {\color{darkblue} \textbf{{N-1}}} wrapped with <START> and <END>. Only wrap single image index digits between <START> and <END>.
}
\end{myquote}
\end{center}

\vspace{6pt}
\textbf{Feedback reflection $p_{fb}$:}

\begin{center}
\begin{myquote}
{\footnotesize
You are a helpful assistant.
\newline

You are iteratively refining the sentence prompt by analyzing the images produced by an AI art generation model, seeking to find out the differences between the user imagined IDEA of the scene and the actual output.

If the generated image is not perfect, provide key REASON on ways to improve the image and sentence prompt to better follow the user imagined IDEA of the scene. Here are some rules to write good key REASON:

- Carefully compare the current image with the IDEA to strictly follow the details described in the IDEA, including object counts, attributes, entities, relationships, sizes, and appearance. Write down what is different in detail.

- Avoid hallucinating information or asks that is not mentioned in IDEA.

- Explain images and unusual entities in IDEA with detailed text descriptions of the scene.

- Explain how to modify prompts to address the given reflection reason.

- Focus on one thing to improve in each REASON. 

- Avoid generating REASON identical with the REASON in previous rounds.

IDEA: {\color{darkblue} \textbf{{\idea input}}}.

End of IDEA.

This is the round {\color{darkblue} \textbf{{t}}} of the iteration.

The iteration history are:

{\color{darkblue} \textbf{{Memory module history.}}}

Based on the above information, you will write REASON that is wrapped with <START> and <END>.

REASON:
}
\end{myquote}
\end{center}

\vspace{6pt}
\textbf{Feedback reflection $p_{revise}$:}

\begin{center}
\begin{myquote}
{\footnotesize
You are a helpful assistant.

Instruction: Given a user imagined IDEA of the scene, converting the IDEA into a sentence prompt that will be used to generate an image.

Here are some rules to write good prompts:

- Each prompt should consist of a description of the scene followed by modifiers divided by commas.

- The modifiers should alter the mood, style, lighting, spatial details, and other aspects of the scene.

- Multiple modifiers can be used to provide more specific details.

- When generating prompts, reduce abstract psychological and emotional descriptions.

- When generating prompts, explain images and unusual entities in IDEA with detailed descriptions of the scene.

- Do not mention 'given image' in output, use detailed texts to describe the image in IDEA.

- Generate diverse prompts.

- Output prompt should have less than 50 words.

IDEA: {\color{darkblue} \textbf{{\idea input}}}.

End of IDEA.

You are iteratively improving the sentence prompt by looking at the images generated by an AI art generation model and find out what is different from the given IDEA.

This is the round {\color{darkblue} \textbf{{t}}} of the iteration.

The iteration history are:

{\color{darkblue} \textbf{{Memory module history.}}}

Generated sentence prompt for current round {\color{darkblue} \textbf{{t}}} is: {\color{darkblue} \textbf{{prompt}}}

Corresponding image generated by the AI art generation model: {\color{darkblue} \textbf{{image}}}

However, {\color{darkblue} \textbf{{reflection}}}

Based on the above information, to improve the image, you will write {\color{darkblue} \textbf{{N}}} detailed prompts exactly about the IDEA follow the rules. Make description of the scene more detailed and add modifiers to address the given key reasons to improve the image. Avoid generating prompts identical with the ones in previous rounds. Each prompt is wrapped with <START> and <END>.

}
\end{myquote}
\end{center}

\bibliographystyle{splncs04}
\bibliography{egbib}
\end{document}